\documentclass[10pt,twocolumn,letterpaper]{article}

\usepackage{cvpr}
\usepackage{times}
\usepackage{epsfig}
\usepackage{graphicx}
\usepackage{amsmath}
\usepackage{amssymb}
\usepackage{adjustbox}
\usepackage{multirow}
\usepackage{algorithm, algorithmic}
\usepackage{footmisc}
\usepackage[toc,page]{appendix}

\usepackage[pagebackref=true,breaklinks=true,letterpaper=true,colorlinks,bookmarks=false]{hyperref}
\cvprfinalcopy % *** Uncomment this line for the final submission

\def\cvprPaperID{8} % *** Enter the CVPR Paper ID here

\ifcvprfinal\fi
\newcommand{\minisection}[1]{\vspace{0.04in} \noindent {\bf #1}}

\newcommand{\introfigure}{%
\begin{figure}
	\centering

    \includegraphics[width=\columnwidth]{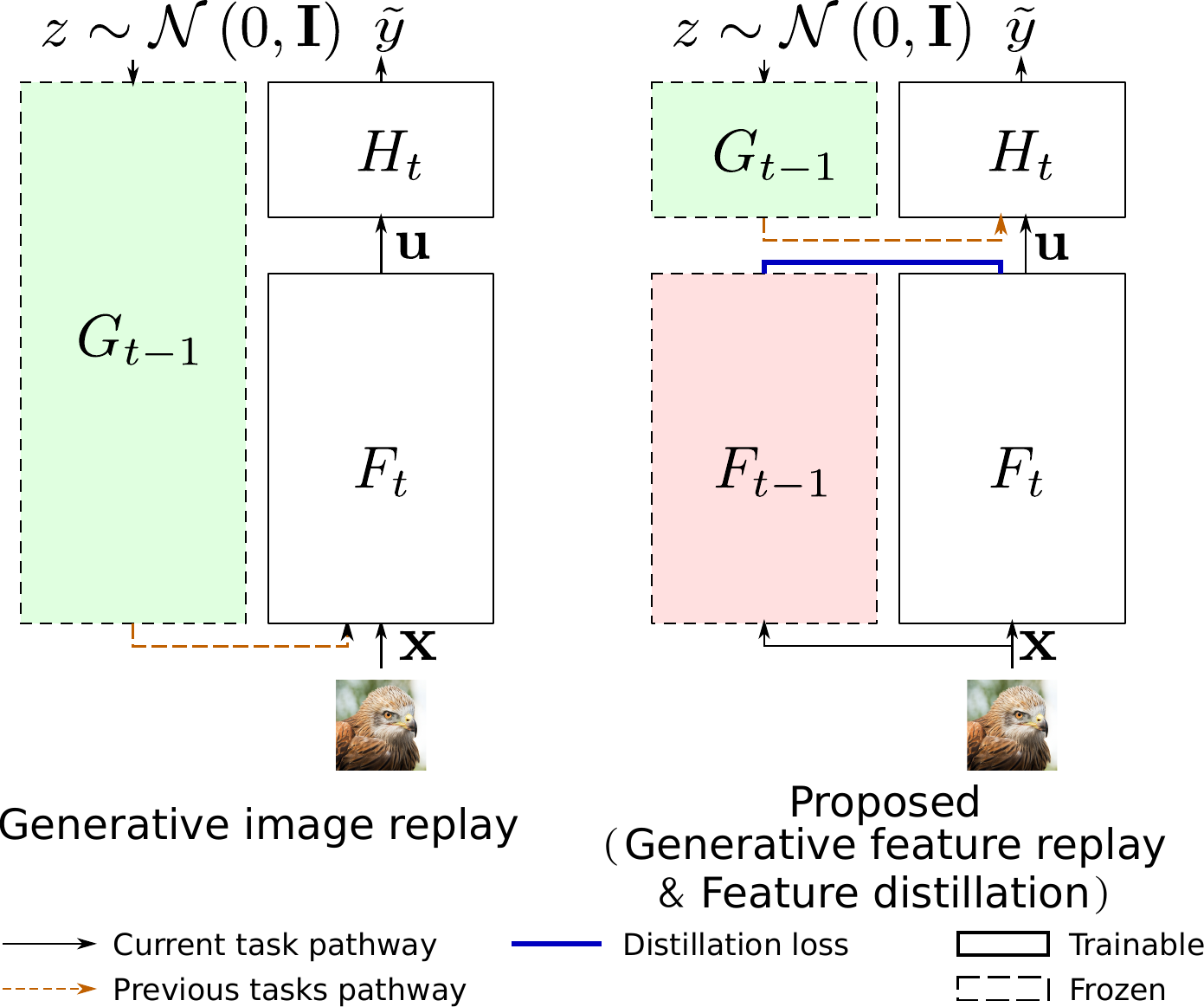}
	\caption{Comparison of generative image replay and the proposed generative feature replay. Instead of replaying images $x$ the proposed method uses a generator $G$ to replay features $u$. To prevent forgetting in the feature extractor $F$ we apply feature distillation. Feature replay allows us to train classifiers $H$ which do not suffer from the imbalance problem common to class-incremental methods. Furthermore, feature generation is significantly easier than image generation and can be applied to complex datasets. }
	\label{fig:motivation}
\end{figure}
}

\newcommand{\imagenetsubfigure}{%
\begin{figure*}
	\centering
	\includegraphics[width=0.33\textwidth]{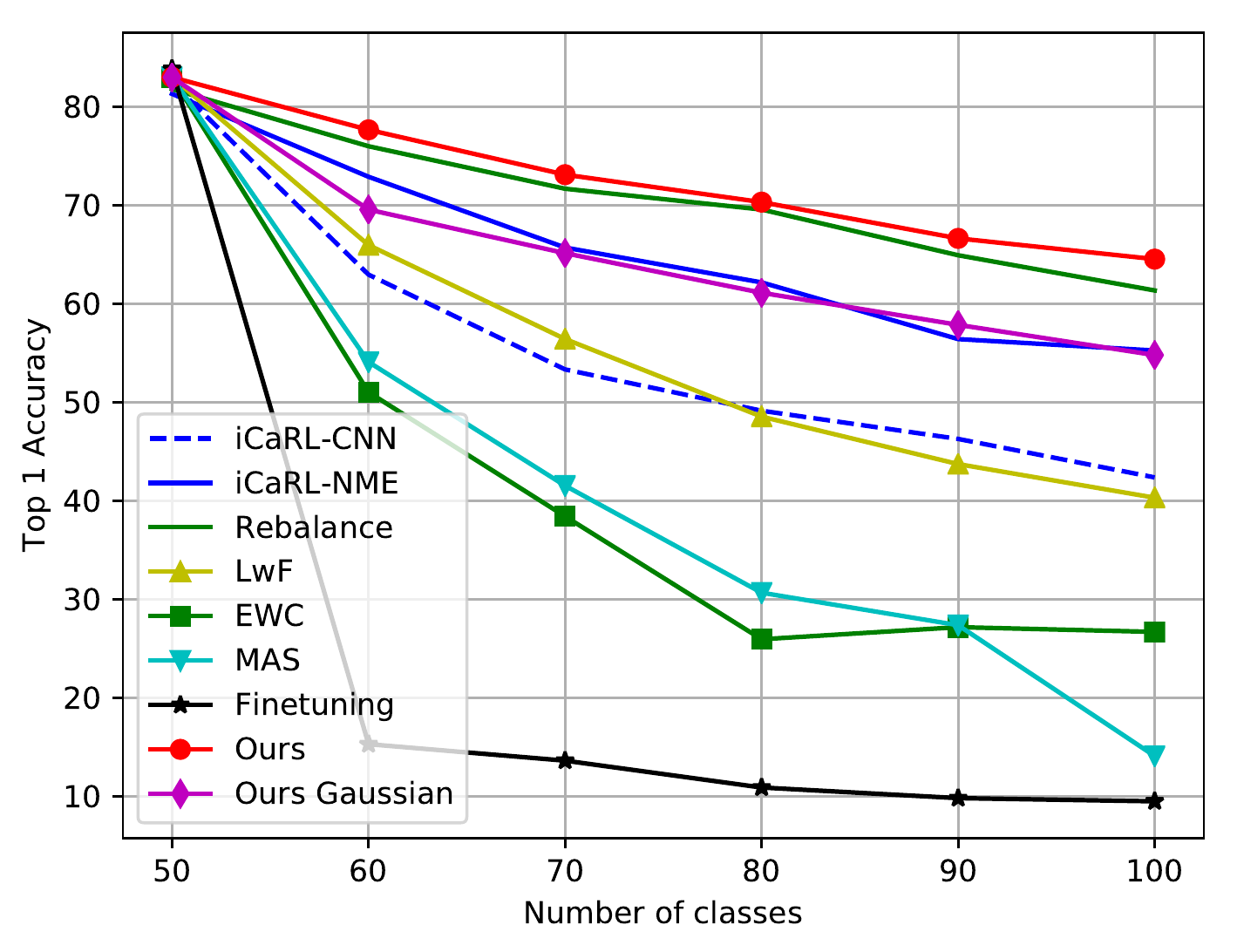}
	\includegraphics[width=0.33\textwidth]{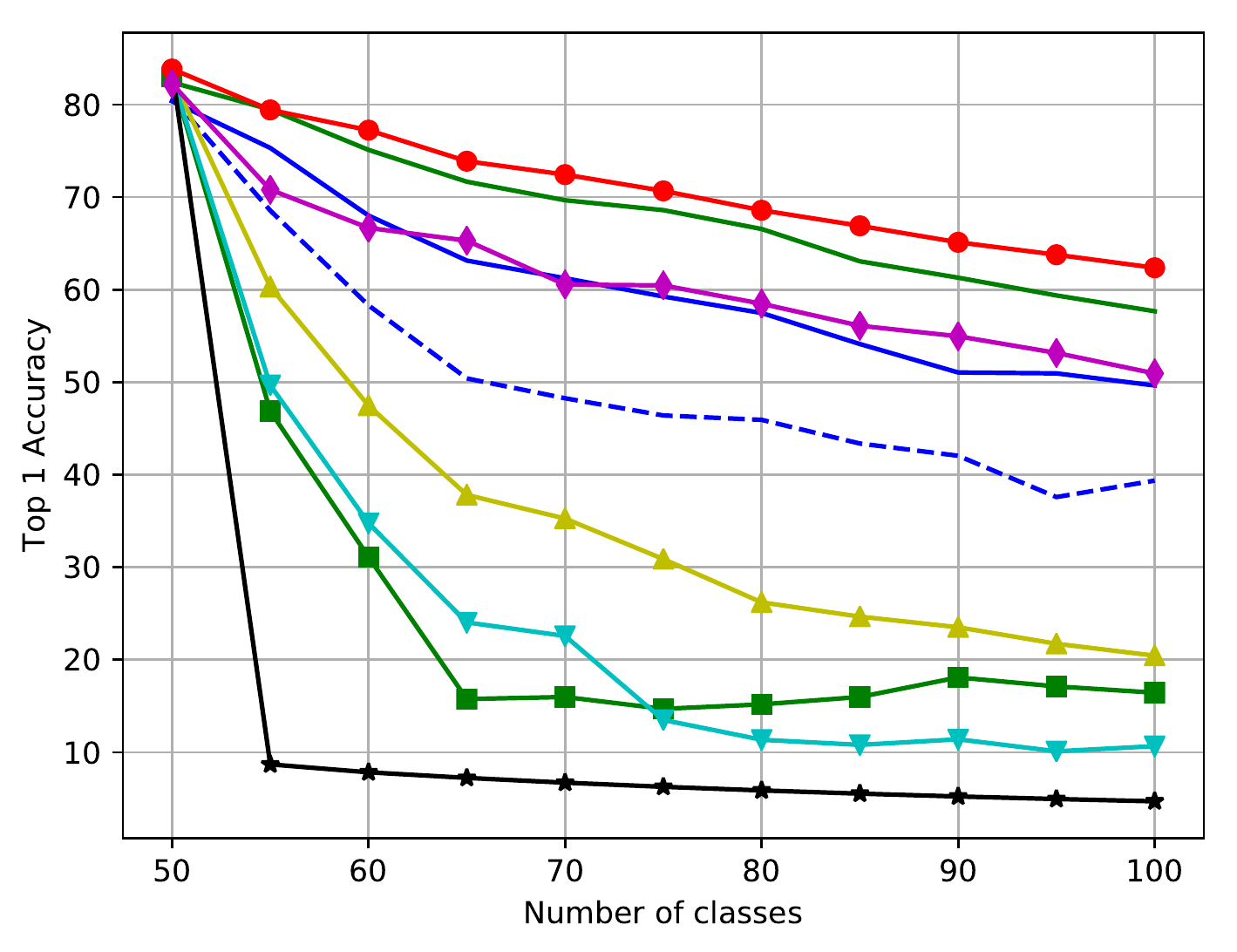}
	\includegraphics[width=0.33\textwidth]{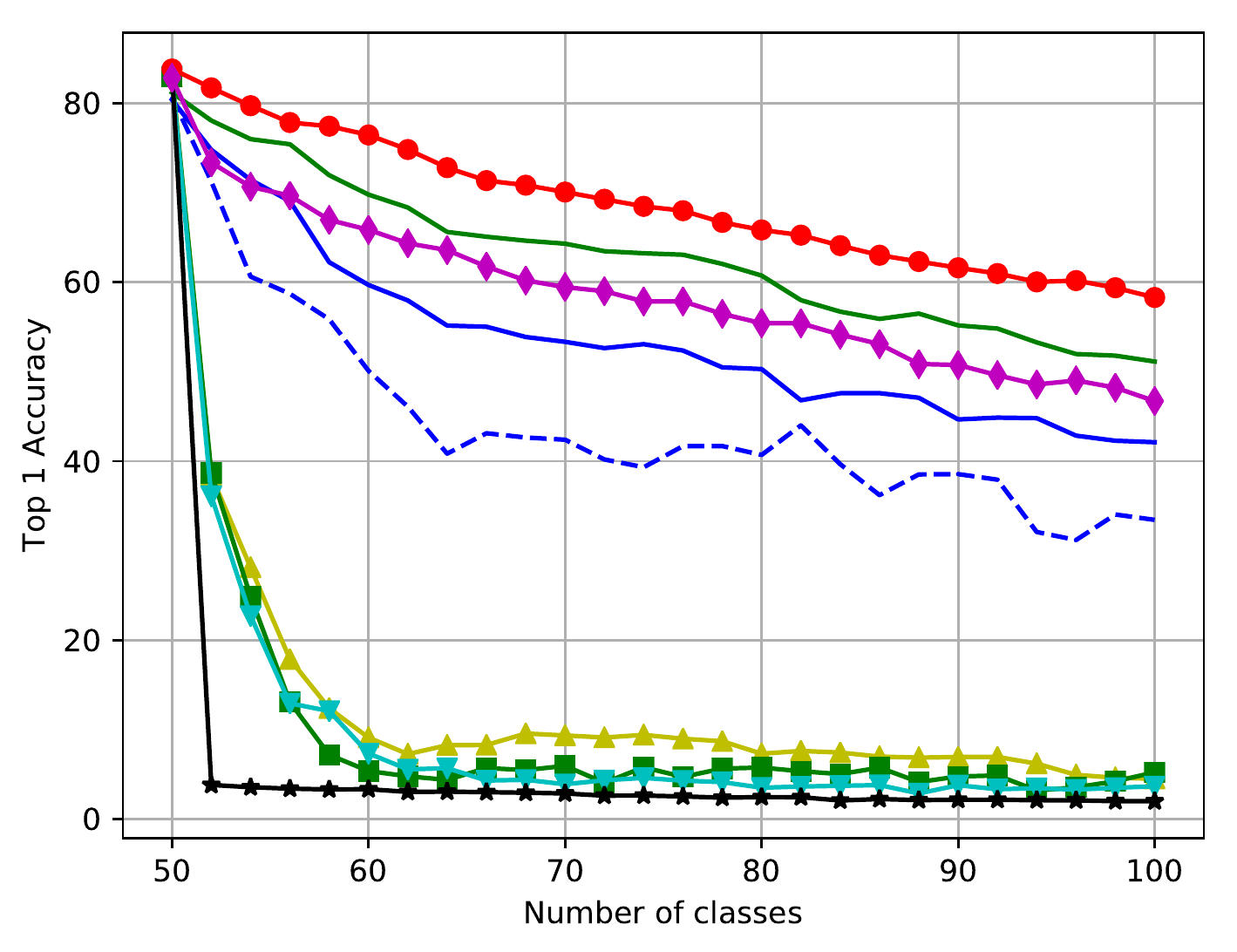}
	\includegraphics[width=0.33\textwidth]{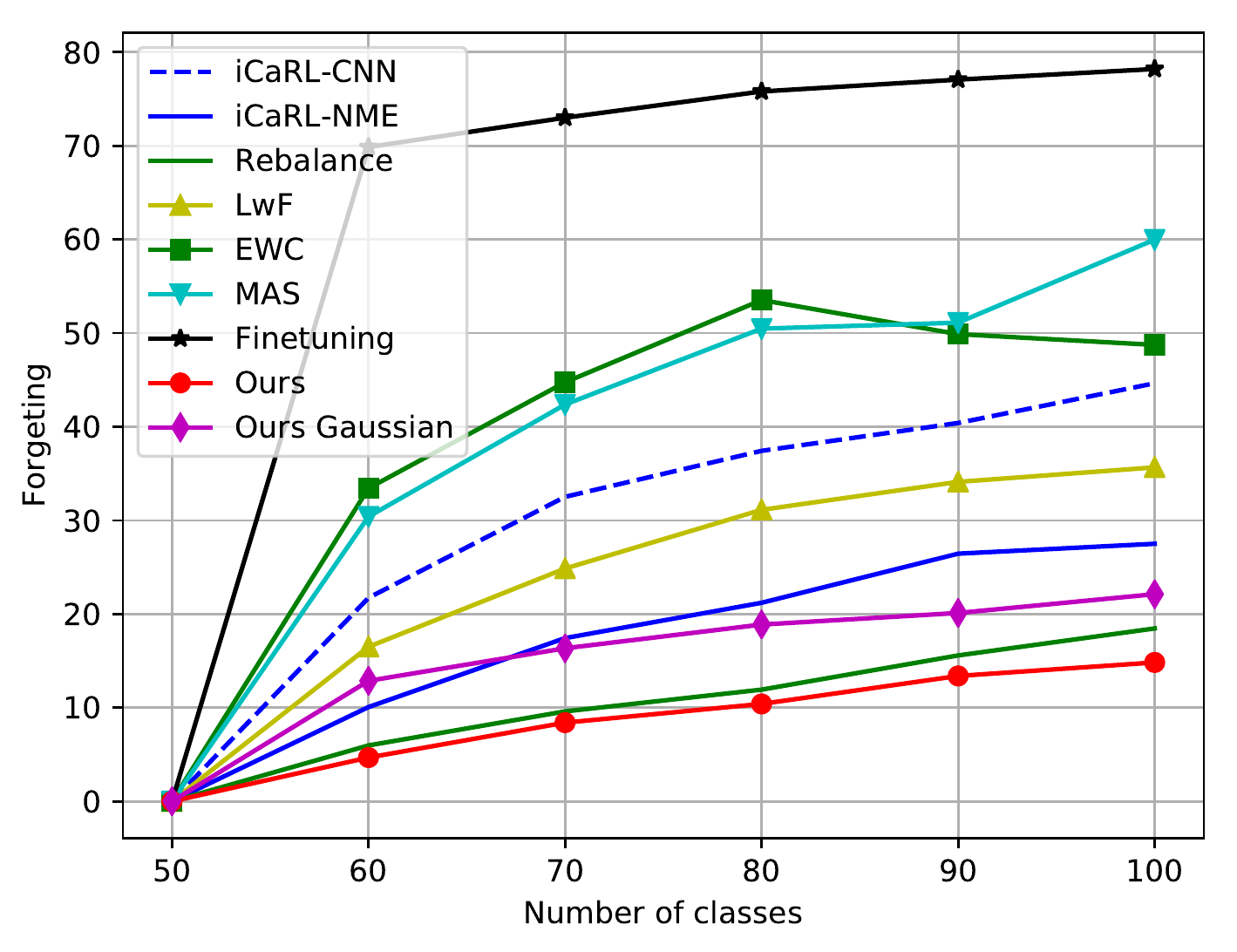}
	\includegraphics[width=0.33\textwidth]{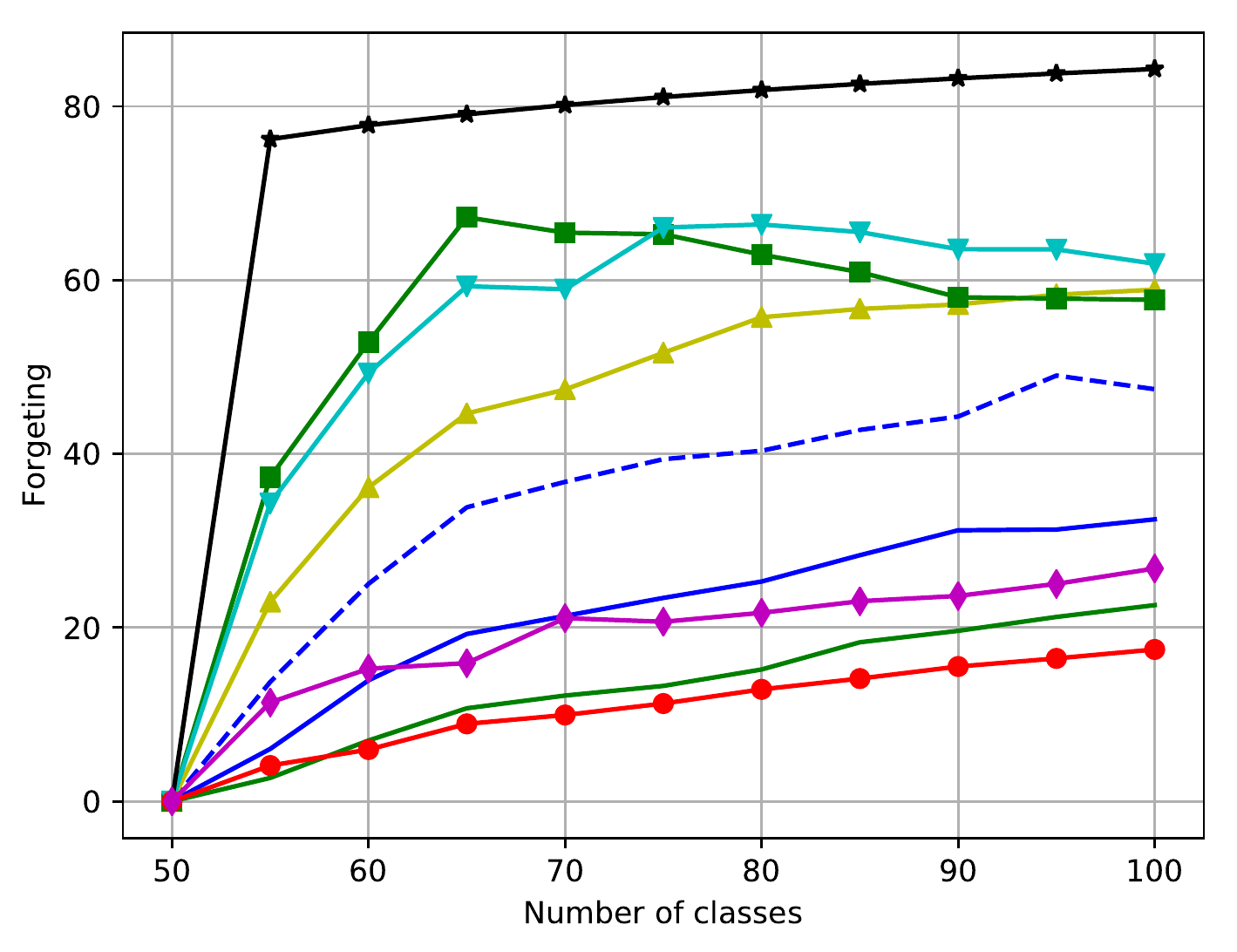}
	\includegraphics[width=0.33\textwidth]{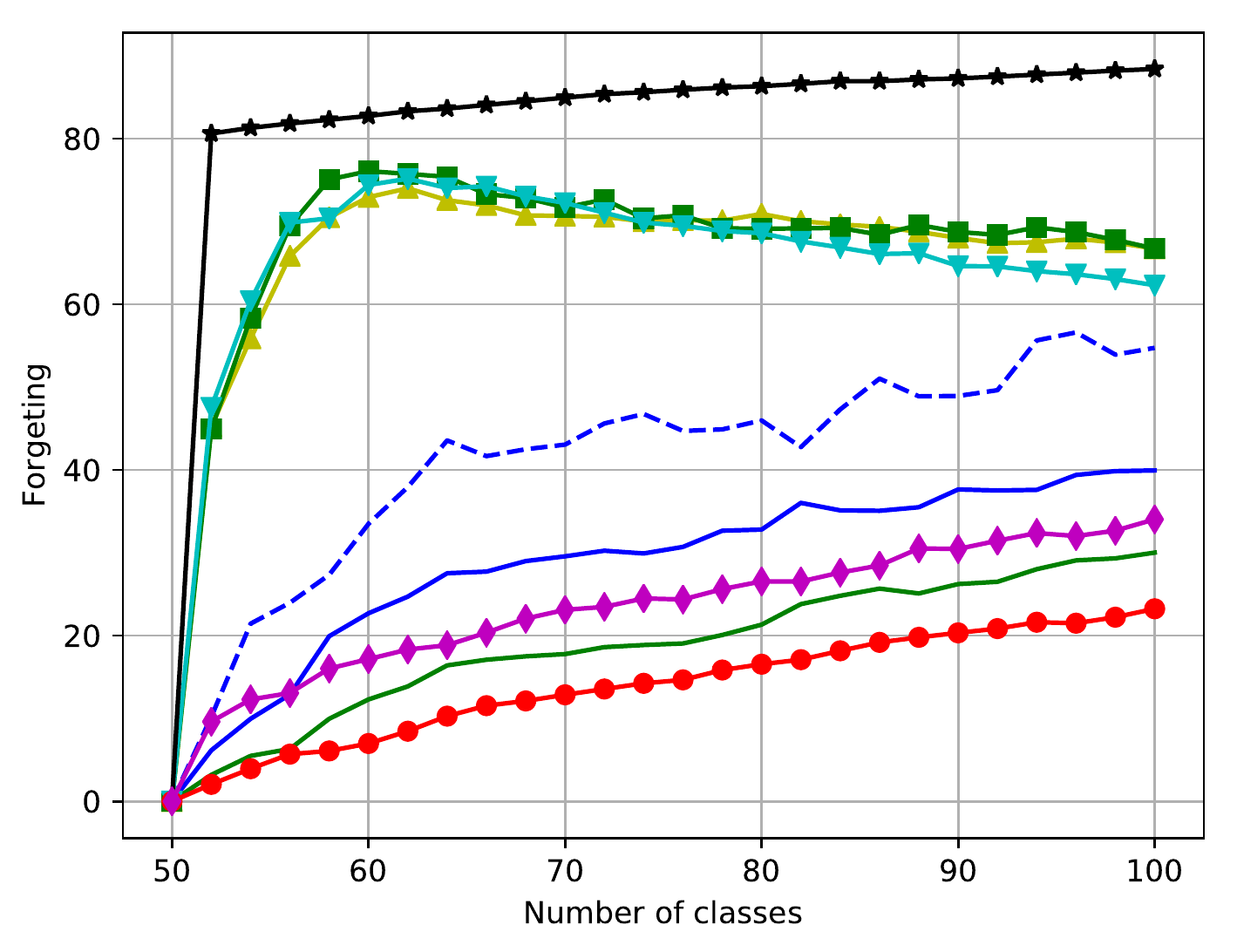}
	\caption{Comparison in the average accuracy (Top) and the average forgetting (Bottom) with various methods on ImageNet-Subset. The first task has the half number of classes, and the remaining classes are divided into 5, 10, 25 tasks respectively. The lines with symbols are methods without using any exemplars, and without symbols are methods with 2000 exemplars. (Joint Training: 77.6)}
	
	\label{fig:imagenet100_50}
\end{figure*}
}

\newcommand{\cifarfigure}{%
\begin{figure*}
	\centering
	\includegraphics[width=0.33\textwidth]{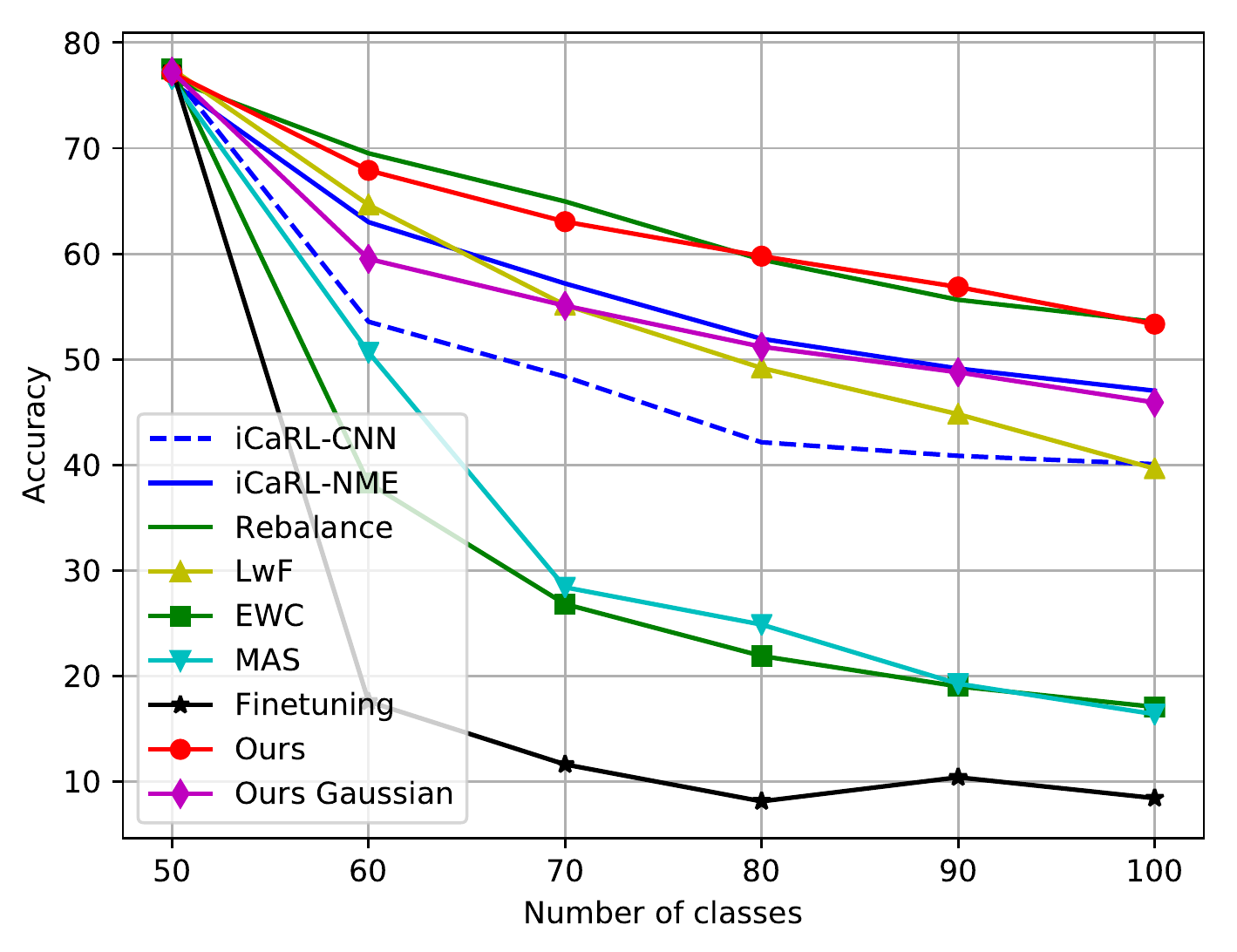}
	\includegraphics[width=0.33\textwidth]{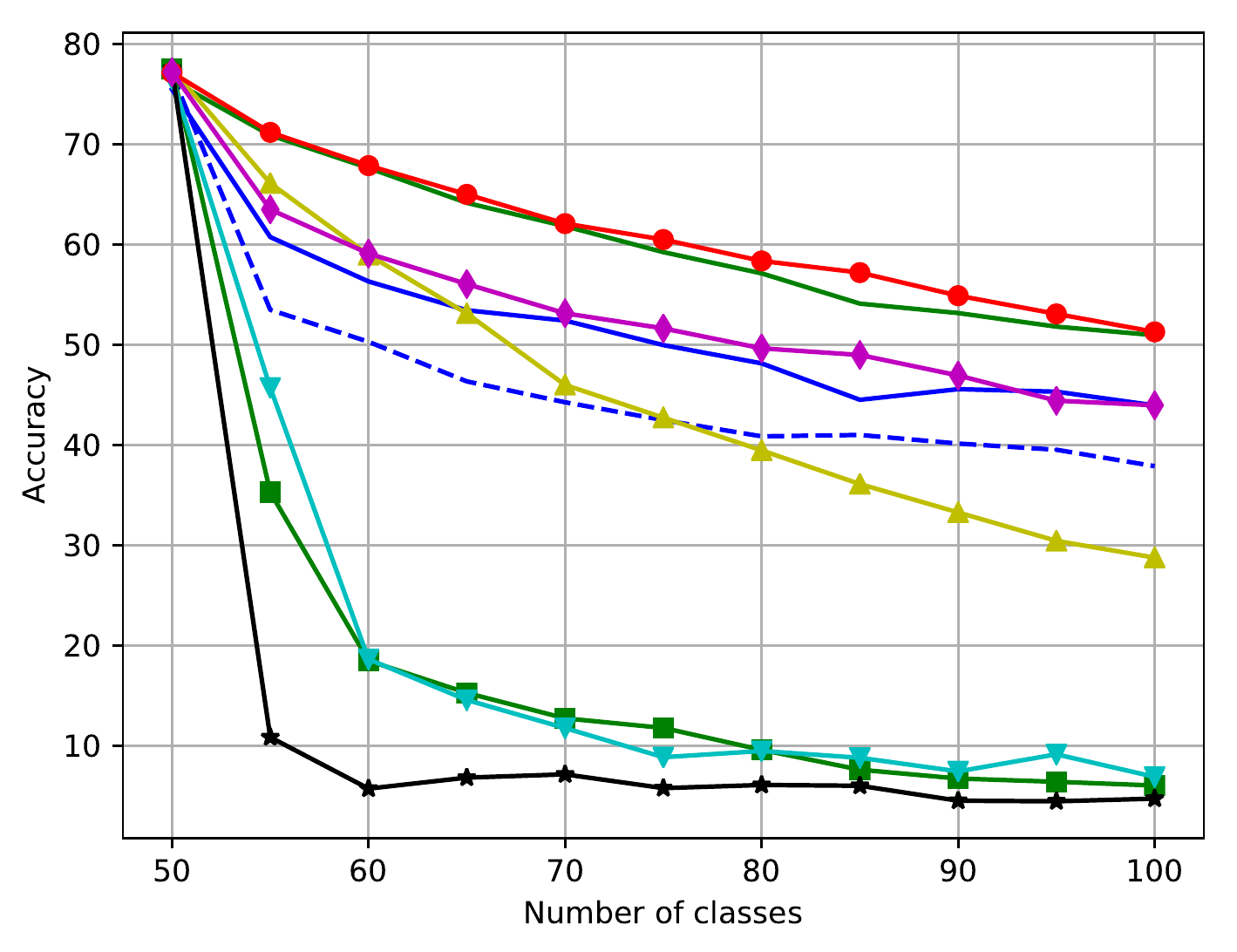}
	\includegraphics[width=0.33\textwidth]{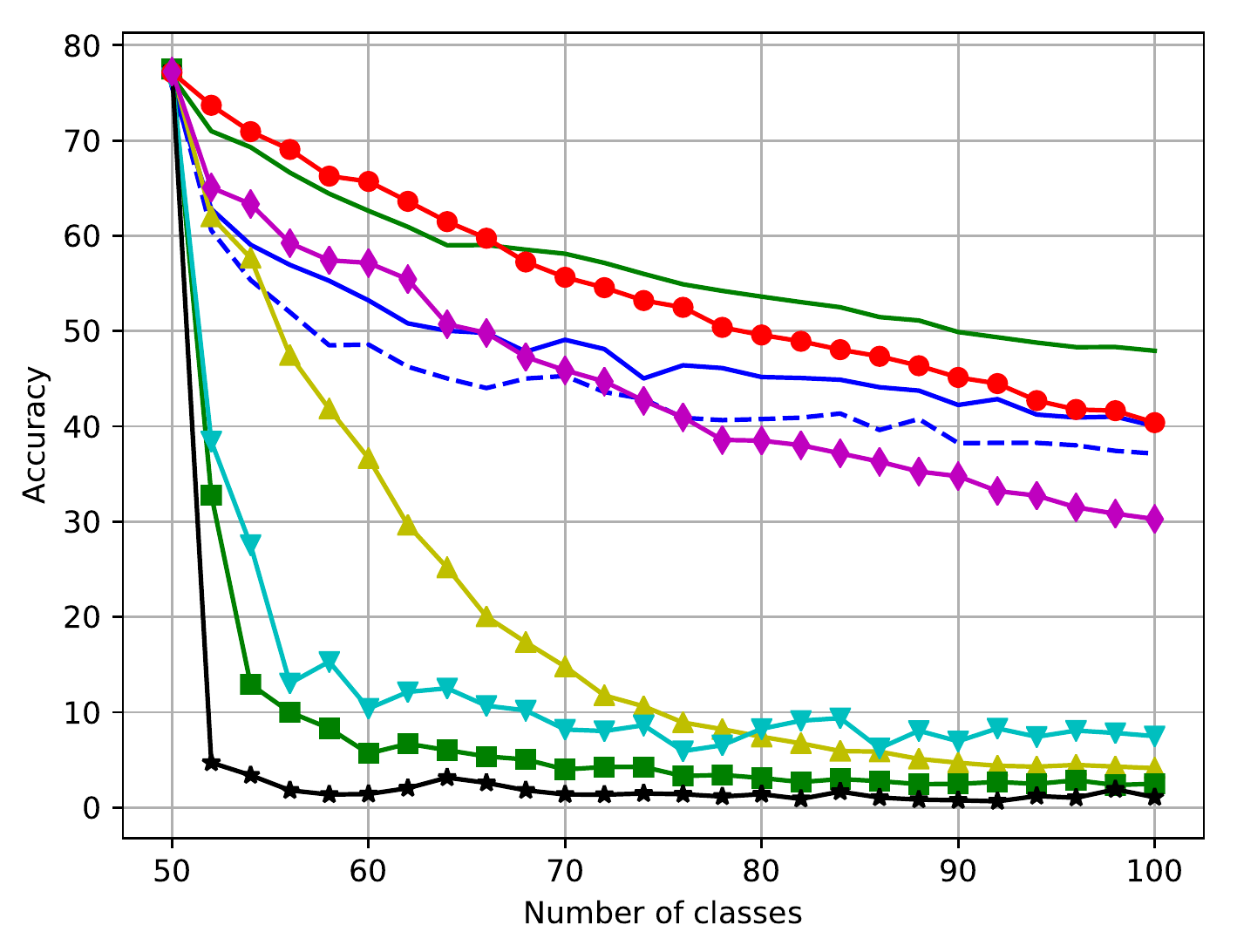}
	\includegraphics[width=0.33\textwidth]{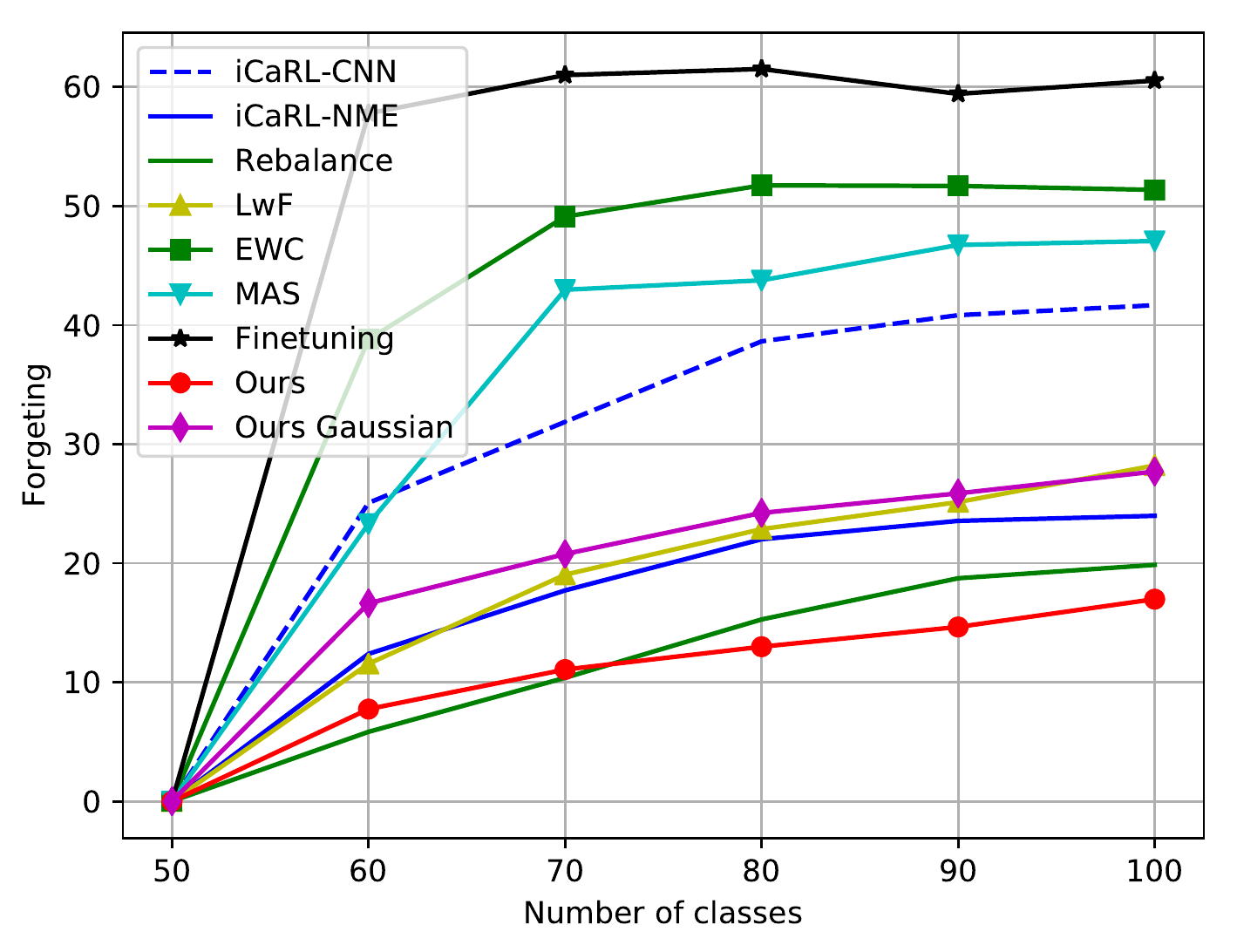}
	\includegraphics[width=0.33\textwidth]{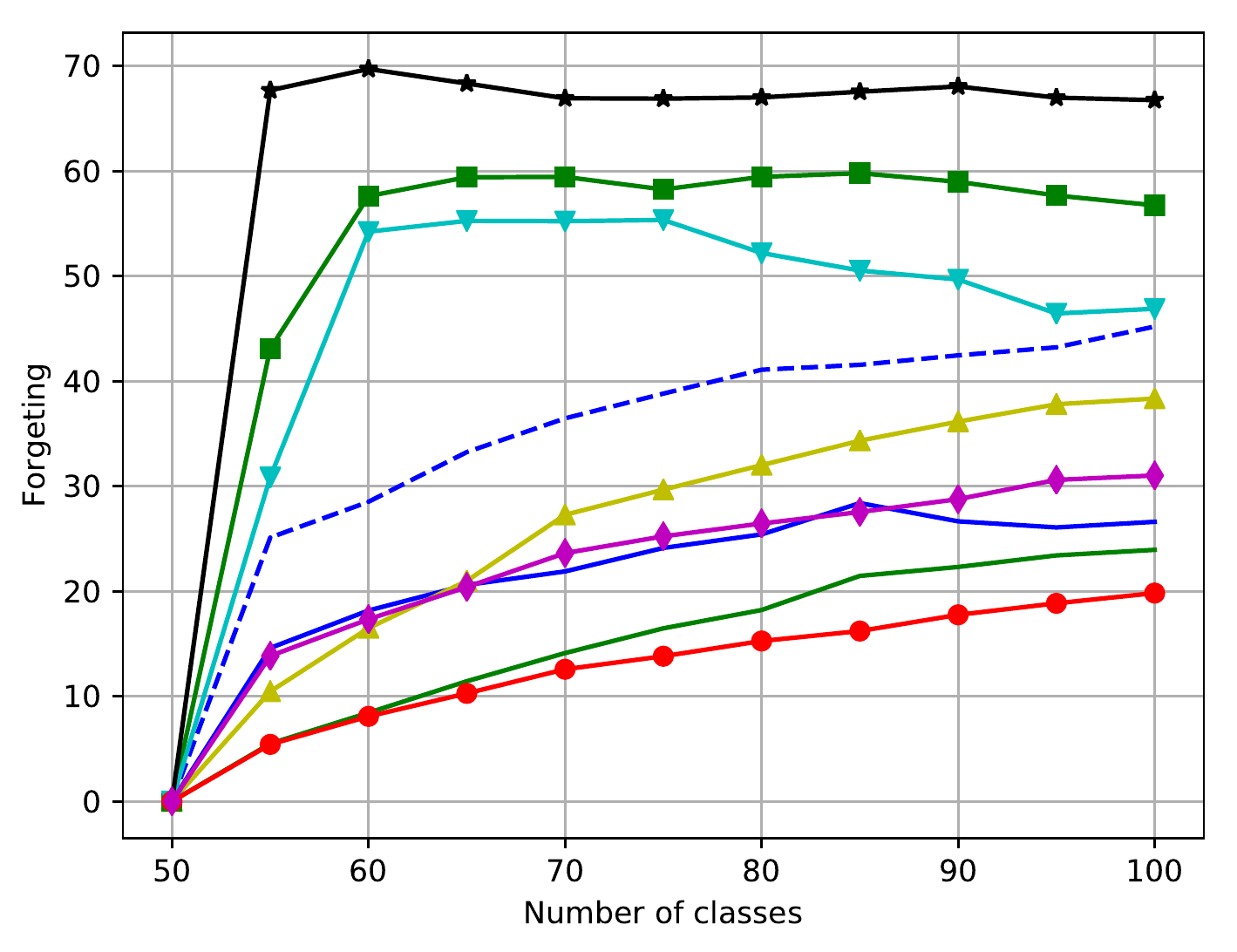}
	\includegraphics[width=0.33\textwidth]{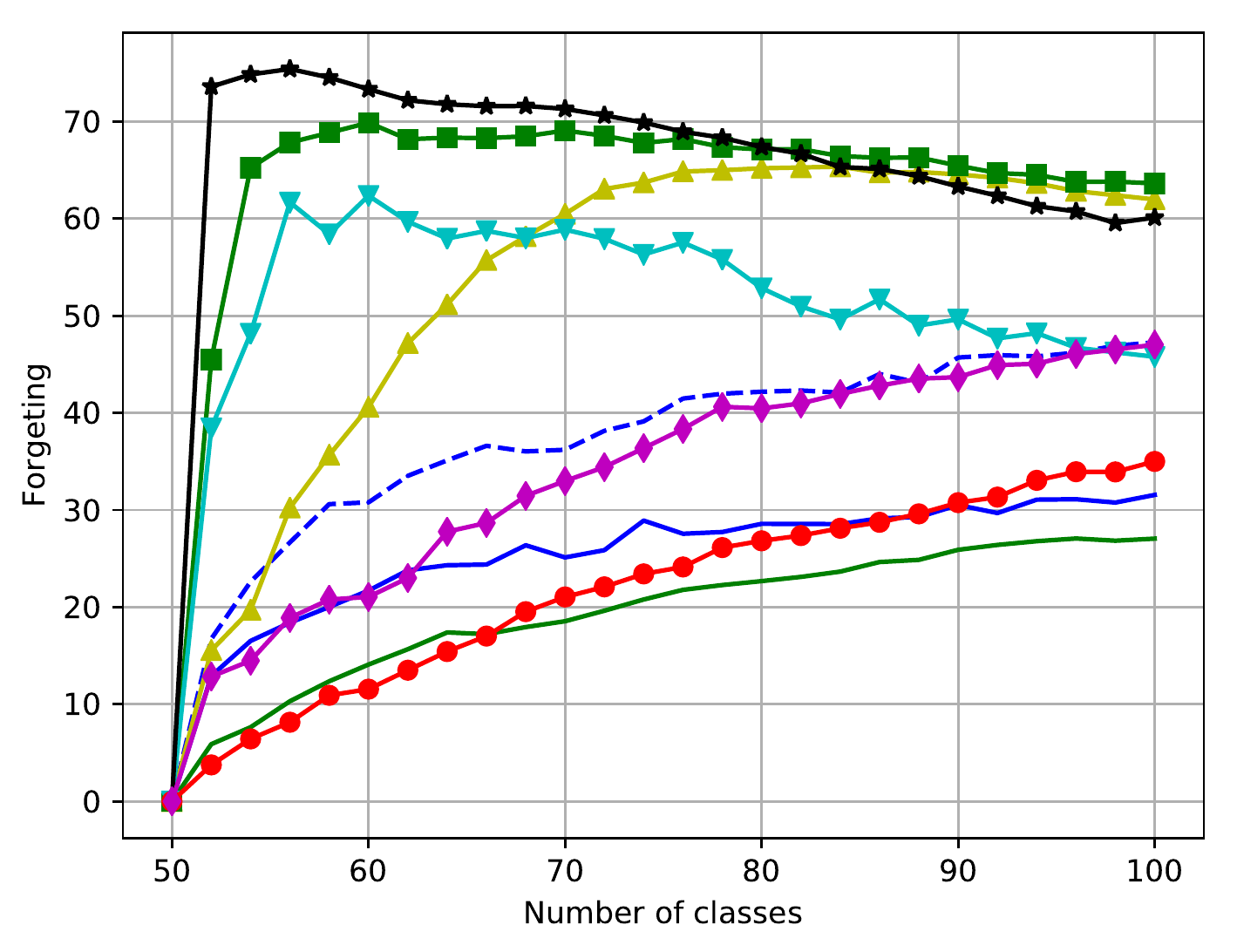}
	\caption{Comparison in the average accuracy (Top) and the average forgetting (Bottom) with various methods on CIFAR-100. The lines with symbols are methods without using any exemplars, and without symbols are methods with 2000 exemplars. (Joint Training: 72.0)}
	
	\label{fig:cifar100_50}
\end{figure*}
}

\newcommand{\ccafigure}{%
\begin{figure*}
	\centering
	\includegraphics[width=\textwidth]{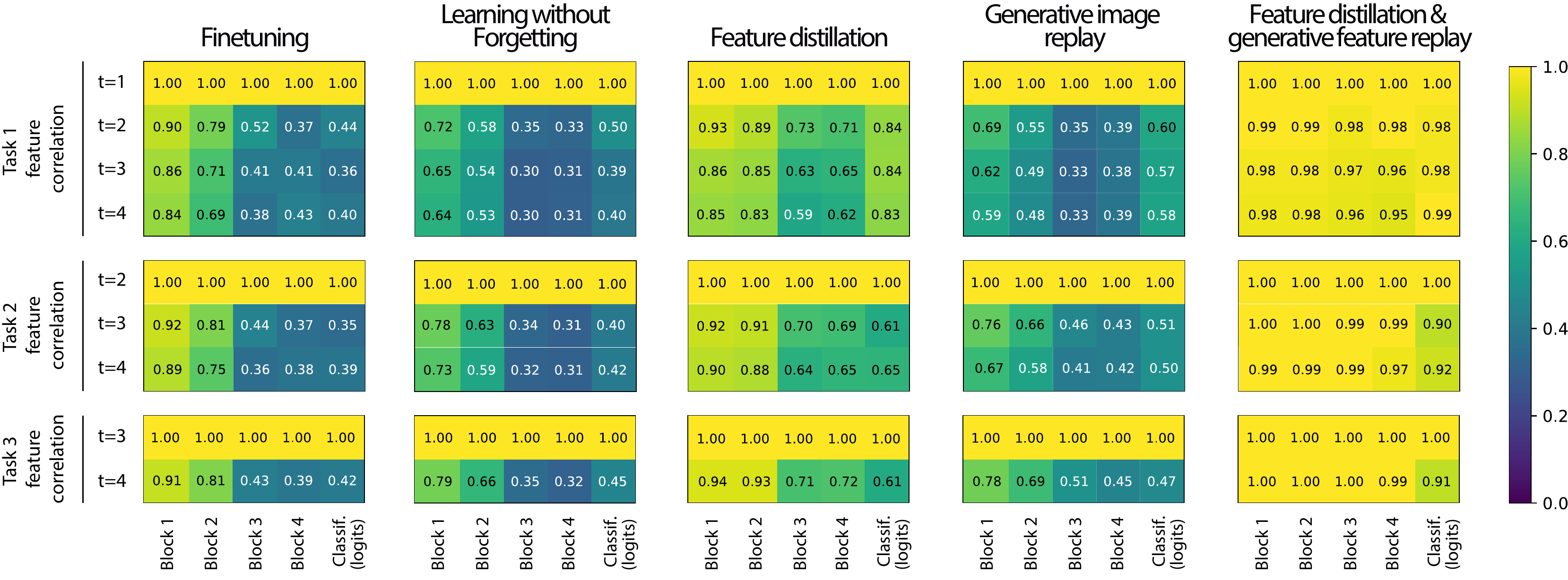}
	\caption{Canonical Correlation Analysis (CCA) similarity of different continual learning methods performed on equally distributed 4-task scenario on CIFAR-100. The vertical axis shows the evolution over time of the correlation for given task activations. The horizontal axis shows correlation at different layers of the network.}
	\label{fig:feature_correlation_analysis}
\end{figure*}
}

\newcommand{\frameworkfigure}{%
\begin{figure*}
	\centering
	\includegraphics[width=0.95\textwidth]{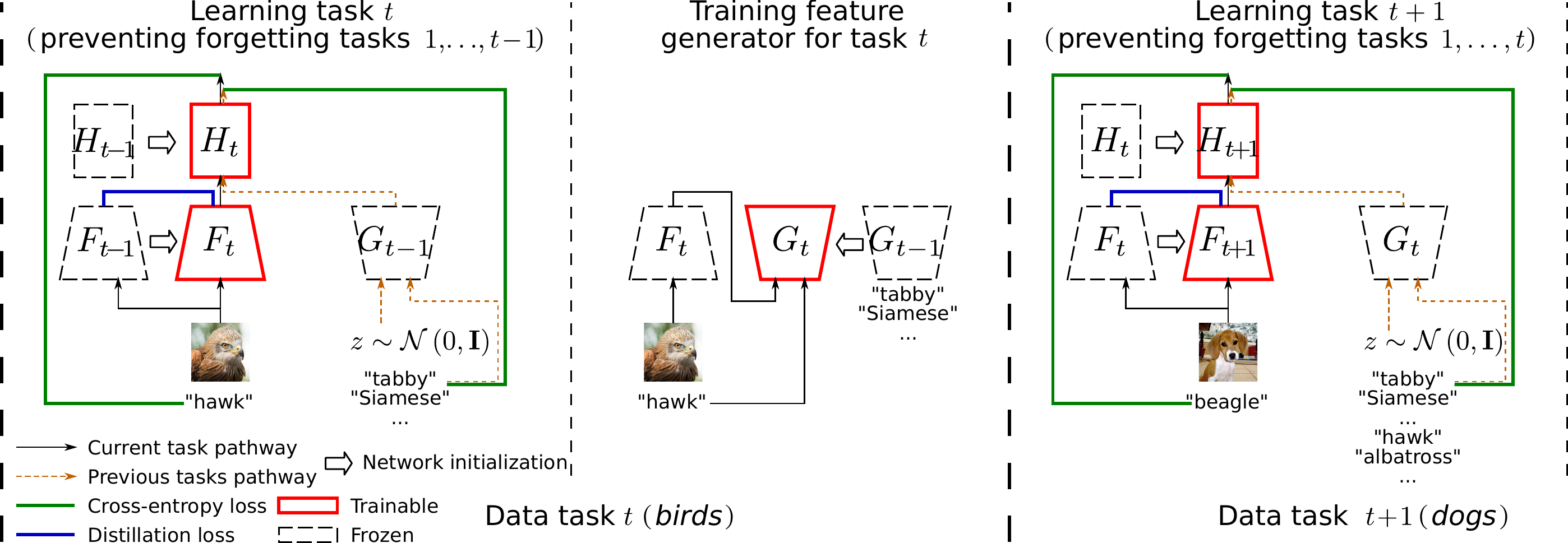}
	\caption{Proposed framework. Distillation and feature generation are used during training to prevent forgetting previous tasks. Once the task is learned, the feature generator is updated with adversarial training and distillation to prevent forgetting in the generator.}
	\label{fig:framework}
\end{figure*}
}

\newcommand{\tsnefigure}{%
\begin{figure*}
	\centering
	\includegraphics[width=0.95\textwidth]{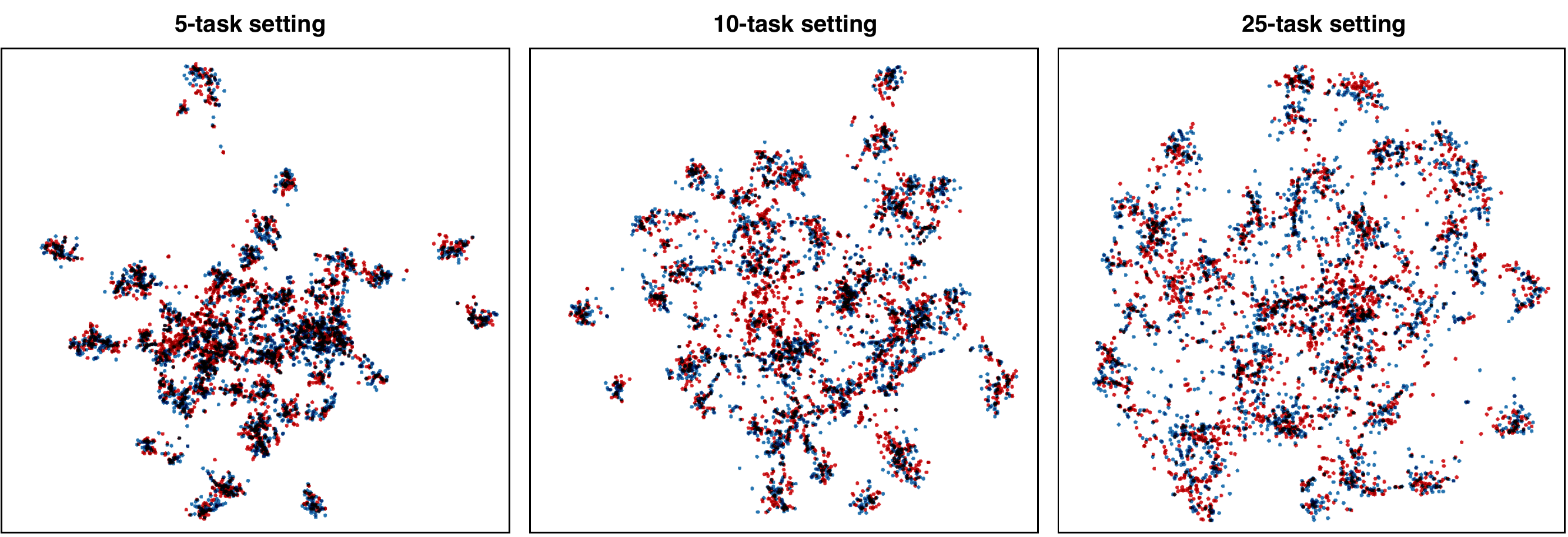}
	\caption{Real features (Red) and Generated features (Blue) on ImageNet-Subset of first task after training all tasks in 5, 10 and 25 tasks setting, respectively.}
	\label{fig:tsne}
\end{figure*}
}

\newcommand{\memorytable}{%
\begin{table}[tb]
\centering
\caption{Memory use comparison between exemplar-based methods, generative image replay (MeRGAN), and Ours.}
\label{tab:memory}
\resizebox{0.48\textwidth}{!}{%
\begin{tabular}{llllll}
\hline
Method & Datasets & Image Size & Exemplar & ResNet-18  & GAN \\
\hline
\multirow{3}{*}{Exemplar-based} & CIFAR-100 & 32x32x3 & 2000 (6.2 Mb) & 42.8 Mb $^*$ & --\\ 
 & ImageNet-100 & 256x256x3 & 2000 (375 Mb) & 45 Mb & -- \\ 
 & ImageNet-1000 & 256x256x3 & 20000 (3.8 Gb) & 45 Mb & -- \\ \hline
MeRGAN & -- & -- & -- & 45 Mb & 8.5 Mb \\
Ours & -- & -- & -- & 45 Mb & 4.5 Mb \\
\hline
\end{tabular}}
\end{table}
}

\newcommand{\ablationFeatureReplay}{%
\begin{table}[tb]
\centering
\caption{Ablation study of replaying different features on CIFAR-100 for the 4-task scenario. For generative image replay, we use MeRGAN~\cite{wu2018memory}, Blocks 1, 2, and 3 are the features after the corresponding residual block in ResNet. Block 4 is the high-level linear features for our method. Average accuracy of all tasks is reported.}
\label{tab:cifar-feat}
\begin{adjustbox}{max width=0.95\textwidth}
\begin{tabular}{lclll}
\hline
                             & \multicolumn{1}{l}{T1}                     & T2   & T3   & T4              \\ \hline
\multicolumn{1}{l|}{Image (MeRGAN)}   & \multicolumn{1}{l|}{82.4}                  & 37.7 & 17.8 & 9.7            \\ \cline{2-2}
\multicolumn{1}{l|}{Block 1} & \multicolumn{1}{c|}{\multirow{4}{*}{80.7}} & 41.6 & 26.5 & 20.1          \\
\multicolumn{1}{l|}{Block 2} & \multicolumn{1}{c|}{}                      & 41.0 & 26.5 & 20.0         \\
\multicolumn{1}{l|}{Block 3} & \multicolumn{1}{c|}{}                      & 51.1 & 37.0 & 26.6         \\
\multicolumn{1}{l|}{Block 4 (Ours)} & \multicolumn{1}{c|}{}                      & 57.6 & 48.2 & 41.5  \\ \hline
\end{tabular}
\end{adjustbox}
\end{table}
}

\newcommand{\ablationalignment}{%
\begin{table}[tb]
\centering
\caption{Ablation study of different regularization methods on CIFAR-100 for the 4-task scenario.}
\label{tab:cifar-reg}
\begin{adjustbox}{max width=\linewidth}
\begin{tabular}{l|llll}
\hline
        & T1                                         & T2   & T3   & T4     \\ \hline
EWC + GAN       & \multicolumn{1}{l|}{\multirow{3}{*}{81.9}} & 40.8 & 26.8 & 21.2 \\ 
MAS + GAN       & \multicolumn{1}{l|}{}                      & 40.2 & 26.0 & 20.9  \\  
Feature Distillation + GAN & \multicolumn{1}{l|}{}                      & 58.4 & 48.8 & 42.2  \\ \hline
\end{tabular}
\end{adjustbox}
\end{table}
}

\newcommand{\tablealgorithm}{%
\begin{algorithm}[tb]
\caption{: Class-incremental task learning. }
\label{table:algorithm}
\begin{algorithmic}
\STATE\textbf{Input:} Sequence $\mathcal{D}_1, \ldots ,  \mathcal{D}_T$, where $\mathcal{D}_t = \left(\mathcal{X}_t,\mathcal{C}_t\right)$. \\
\STATE\textbf{Require:} Feature extractor $F_0$, Classifier $H_0$,\\
\ \ \ \ \ \ \ \ \ \ \ \ \ \ \ \ Generator $G_0$. All trained end-to-end. \\
\textbf{for} $t=1, \ldots , T$ \\
\ \ \ \ \textbf{if} $t = 1$ \\
\ \ \ \ \ \ \ \ Step 1: Train $F_1$ and $H_1$ with $ \mathcal{D}_1 $.\\

\ \ \ \ \ \ \ \ Step 2: Train  $G_1$ with $\mathbf{u}_1 = F_1(\mathbf{x}_i), \forall x_i \in \mathcal{D}_1 $. \\

\ \ \ \ \textbf{else} \\
\ \ \ \ \ \ \ \ Step 3: Train $F_t$ and $H_t$ with $ \mathcal{D}_t $ and generated\\
\ \ \ \ \ \ \ \ \ \ \ \ \ \ \ \ \ \ \ \ \ features $\hat{\mathbf{u}}_{t^\prime} = G_{t-1}(\mathcal{C}_{t^\prime}, \mathbf{z})$, where $\mathcal{C}_{t^\prime}$  is \\ 
\ \ \ \ \ \ \ \ \ \ \ \ \ \ \ \ \ \ \ \ \ all previous classes.\\

\ \ \ \ \ \ \ \ Step 4: Train $G_t$ with $\mathbf{u}_t = F_t(\mathbf{x}_i), \forall x_i \in \mathcal{D}_1$ and\\
\ \ \ \ \ \ \ \ \ \ \ \ \ \ \ \ \ \ \ \ \ $\hat{\mathbf{u}}_{t^\prime} = G_{t-1}(\mathcal{C}_{t^\prime}, \mathbf{z})$\\
\textbf{end for}
\end{algorithmic}

\end{algorithm}
}

\begin{document}

%%%%%%%%% TITLE
\title{Generative Feature Replay For Class-Incremental Learning}

\author{Xialei Liu$^{1, *}$, Chenshen Wu$^{1, *}$, Mikel Menta$^{1}$, Luis Herranz$^{1}$, Bogdan Raducanu$^{1}$,  \\
Andrew D. Bagdanov$^{2}$, Shangling Jui$^{3}$, Joost van de Weijer$^{1}$\\
$^{1}$ Computer Vision Center, Universitat Autonoma de Barcelona, Barcelona, Spain\\
$^{2}$ Media Integration and Communication
Center, University of Florence, Florence, Italy \\
$^{3}$ Huawei Kirin Solution, Shanghai, China\\
{\tt\small \{xialei,chenshen,mkmenta,lherranz,bogdan,joost\}@cvc.uab.es}, \\\tt\small{andrew.bagdanov@unifi.it
, jui.shangling@huawei.com}
}

\maketitle
%\thispagestyle{empty}

%%%%%%%%% ABSTRACT
\begin{abstract}
Humans are capable of learning new tasks without forgetting previous ones, while neural networks fail due to catastrophic forgetting between new and previously-learned tasks. We consider a class-incremental setting which means that the task-ID is unknown at inference time. The imbalance between old and
 new classes typically results in a bias of the network towards the newest ones. This imbalance problem can either be addressed by storing exemplars from previous tasks, or by using image replay methods. However, the latter can only be applied to toy datasets since image generation for complex datasets is a hard problem. 
  
 We propose a solution to the imbalance problem based on generative feature replay which does not require any exemplars. To do this, we split the network into two parts: a feature extractor and a classifier. To prevent forgetting, we combine generative feature replay in the classifier with feature distillation in the feature extractor. Through feature generation, our method reduces the complexity of generative replay and prevents the imbalance problem. Our approach is computationally efficient and scalable to large datasets. Experiments confirm that our approach achieves state-of-the-art results on CIFAR-100 and ImageNet, while requiring only a fraction of the storage needed for exemplar-based continual learning. Code available at \url{https://github.com/xialeiliu/GFR-IL}. \footnotetext[1]{Both authors contributed equally.}

\end{abstract}

%%%%%%%%%%%%%% TO DO
% joost: I am not convinced with the itemized contribution list..
% joost: I removed {Evaluation protocol} which was not that useful, maybe if we have more info

\section{Introduction}
% CL and catastrophic forgetting
Humans and animals are capable of continually acquiring and updating knowledge throughout their lifetime. The ability to accommodate new knowledge while retaining previously learned knowledge is referred to as \emph{incremental or continual learning}, which is essential to building scalable and reusable artificially intelligent systems. Current deep neural networks have achieved impressive performance on many benchmarks, comparable or even better than humans (e.g. image classification~\cite{he2015delving}). However, when trained for new tasks, these networks almost completely forget the previous ones due to the problem of \textit{catastrophic forgetting}~\cite{mccloskey1989catastrophic} between the new and previously-learned tasks.

%%%%%%%%%%%%%%%%%%%%%%%%%%%%%%%%%%%
%%%%%%%%%%% FIGURE 
%%%%%%%%%%%%%%%%%%%%%%%%%%%%%%%%%%%
\introfigure

To overcome \textit{catastrophic forgetting} several approaches, inspired in part by biological systems, have been proposed. The first category of approaches use regularizers that limit the plasticity of the network  while training on new tasks so the network remains stable on previous ones~\cite{aljundi2018memory,kirkpatrick2017overcoming,li2018learning,liu2018rotate,zenke2017continual}. Another type of approach involves dynamically increasing the capacity of the network to accommodate new tasks~\cite{lee2017lifelong,rusu2016progressive}, often combined with task-dependent masks on the weights~\cite{mallya2018piggyback,mallya2018packnet} or activations~\cite{serra2018overcoming} to reduce the chance of catastrophic forgetting.

A third category of approaches relies on memory replay, i.e. replaying samples of previous tasks while learning with the samples of the current task. These samples could be real ones ('exemplars'), like in~\cite{chaudhry2018riemannian,lopez2017gradient,rebuffi2017icarl} in which we refer to the process as 'rehearsal' or could be synthethic ones obtained through generative mechanisms, in which case we refer to the process as 'pseudo-rehearsal'~\cite{robins1995catastrophic,shin2017continual,wu2018memory}. Incremental learning methods are typically evaluated and designed for a particular testing scenario~\cite{van2019three}. Task-incremental learning considers the case where the task ID is given at inference time~\cite{lopez2017gradient, mallya2018packnet, serra2018overcoming}. Class-incremental learning considers the more difficult scenario in which the task ID is unknown at testing time \cite{hou2019learning,rebuffi2017icarl,wu2019large}.  

Recently, research attention has shifted from task-incremental to class-incremental learning. The main additional challenge, which class-incremental methods have to address, is balancing the different classifier heads. The imbalance problem occurs because during training of the current task there is none or only limited data available from previous tasks, which biases the classifier towards the most recently learned task. Various solutions to this problem have been proposed. iCarL\cite{rebuffi2017icarl} stores a fixed budget of exemplars from previous tasks in a way that exemplars approximate the mean of classes in the feature space. The nearest-mean classifier is used for inference. Wu et al.~\cite{wu2019large} found that the last fully-connected layer has a strong bias towards new classes, and corrected the bias with a linear model estimated from exemplars. Hou et al.~\cite{hou2019learning}
replace the softmax with a cosine similarity-based loss, which, combined with exemplars, addresses the imbalance problem. All these methods have in common that they require storage of exemplars. However, for many applications -- especially due to privacy concerns or storage restrictions -- it is not possible to store \emph{any} exemplars from previous tasks. 

The only methods which successfully addresses the imbalance problem without requiring any exemplars are methods performing generative replay~\cite{shin2017continual,wu2018memory}. These methods train a generator continuously to generate samples of previous tasks, and therefore prevent the imbalance problem. Thus, these methods report excellent results for class-incremental learning. However, they have one major drawback: the generator should accurately generate images from previous task distributions. For small data sets like MNIST and CIFAR-10 this is feasible, however, for larger datasets with more classes and larger images (like CIFAR-100 and ImageNet) these methods yield unsatisfactory results.   

In this paper, we propose a novel approach based on generative feature replay to overcome catastrophic forgetting in class-incremental continual learning. Our approach is motivated by the fact that image generation is a complex process when the number of images is limited or the number of classes is high. Therefore, instead of image generation we adopt feature generation which is considerably easier than accurately generating images. We split networks into two parts: a feature extractor and a classifier. To prevent forgetting in the entire network, we combine generative feature replay (in the classifier) with feature distillation on the feature extractor. To summarize, our contributions are:
\begin{itemize}
    \item  We design a hybrid model for class-incremental learning which combines generative feature replay at the classifier level and distillation in the feature extractor.
    \item  We provide visualization and analysis based on Canonical Correlation Analysis (CCA) of how and where networks forget in order to offer better insight.
    \item  We outperform other methods which do not use exemplars by a large margin on the ImageNet and CIFAR-100 datasets. Notably, we also outperform methods using exemplars for most of the evaluated settings. Additionally, we show that our method is computationally efficient and scalable to large datasets.
\end{itemize}

\section{Related Work}
\label{sec:related}

\subsection{Continual learning}

Continual learning can be divided into three main categories as follows (more details in the surveys~\cite{de2019continual,parisi2019continual}):

\minisection{Regularization-based methods.}
A first family of techniques is based on regularization. They  estimate the relevance of each network parameter and penalize those parameters which show significant change when switching from one task to another. The difference between these methods lies on how the penalization is computed. For instance, the EWC approach in~\cite{kirkpatrick2017overcoming,liu2018rotate}, weights network parameters using an approximation of the diagonal of the Fisher Information Matrix (FIM). In~\cite{zenke2017continual}, the importance weights are computed online. They keep track of how much the loss changes due to a change in a specific parameter and accumulate this information
during training. A similar approach is followed in \cite{aljundi2018memory}, but here, instead of considering the changes in the loss, they focus on the changes on activations. This way, parameter relevance is learned in an unsupervised manner. Instead of regularizing weights, ~\cite{jung2018less,li2018learning} align the predictions using the data from the current task.

\minisection{Architecture-based methods.}
A second family of methods to prevent catastrophic forgetting produce modifications in a network's morphology by growing a sub-network for each task, either logically or physically~\cite{lee2017lifelong,rusu2016progressive}. Piggyback~\cite{mallya2018piggyback} and Packnet~\cite{mallya2018packnet} and  learn a separate mask for each task, while HAT~\cite{serra2018overcoming} and Ternary Feature Masks~\cite{masana2020ternary} learn a mask on the activations instead of for each parameter.

\minisection{Rehearsal-based methods.}
The third and last family of methods to prevent catastrophic forgetting are rehearsal-based. Existing approaches use two strategies: either store a small number of training samples from previous tasks or use a generative mechanism to sample synthetic data from previously learned distributions. In the first category, iCaRL~\cite{rebuffi2017icarl} stores a subset of real data (called \emph{exemplars}).  For a given memory budget, the number of exemplars stored should decrease when the number of classes increases, which inevitably leads to a decrease in performance. A similar approach is pursued in~\cite{lopez2017gradient}, but the gradients of previous tasks are preserved. An improved version of this approach overcomes some efficiency issues~\cite{chaudry2019AGEM}. 
In~\cite{hou2019learning} the authors propose two losses called the less-forget constraint and inter-class separation to prevent forgetting. The less-forget constraint minimizes the cosine distance between the features extracted by the original and new models. The inter-class separation separates the old classes from the new ones with the stored exemplars used as anchors. In~\cite{wu2019large,belouadah2019il2m}, a bias correction layer to correct the output of the original fully-connected layer is introduced to address the data imbalance between the old and new categories. In~\cite{pellegrini2019latent}, they propose to store activations for replay and a slow-down
learning at all the layers below the replay layer.

Methods in the second category do not store any exemplars, but introduce a generative mechanism to sample data from. In~\cite{shin2017continual}, memory replay is implemented with an unconditional GAN, where an auxiliary classifier is required in order to determine which class the generated samples belong to. An improved version of this approach was introduced in~\cite{wu2018memory}, where they use a class-conditional GAN to generate synthetic data. In contrast, FearNet~\cite{kemker2017fearnet} uses a generative autoencoder for memory replay and ~\cite{xiang2019incremental} generates intermediate features. Using the class statistics from the encoder, synthetic data for previous tasks is generated based on the mean and covariance matrix. The main limitation of this approach is the assumption of a Gaussian distribution of the data and the reliance on pretrained models. 

\subsection{Generative adversarial networks}

Generative adversarial networks (GANs)~\cite{goodfellow2014generative} are able to generate realistic and sharp images conditioned on object categories \cite{grinblat2017class,perarnau2016invertible}, text \cite{reed2016text2image,zhang2017text2image}, another image (image translation) \cite{kim2017image2image,zhu2017image2image} and style transfer \cite{dumoulin2017artistic}. In the context of continual learning, they were successfully been used for memory replay, by generating synthetic samples from previous tasks \cite{wu2018memory}. Here we are going to analyze the GANs limitations and argue why GANs for feature generation are preferable over image generation.

\minisection{Adversarial image generation.}
Although GANs achieved impressive performance recently, in order to generate high-resolution images \cite{brock2019biggan,Karras2018progressivegan}, they are not immune to common GAN problems such as stability (solutions are available at a high computational costs) and the need for a large training set of real images. Additionally, the generation of high-resolution images does not guarantee that they are able to capture a large enough variety of visual concepts with a good discriminative power \cite{dai2017badgan}.
Only recently, the authors in \cite{lucic2019hifigan} proposed to uses high resolution images. 

However, they are not yet sufficient to generate high quality images for the downstream tasks, for instance training a deep neural network classifier. In the case of few-shot and zero-shot learning, only few samples or no sample are existing to train the GANs, which results in even more challenges to generate useful images. 

\minisection{Adversarial feature generation.}
Recently, feature generation has appeared as an alternative to image generation, especially for the cases of few-shot learning, demonstrating superior performance. In \cite{xian2018feature}, they propose a GAN architecture with a classifier on top of the generator, in order to
generate features that are better suited for classification. The same idea is further improved in \cite{xian2019f}, where they combine a better feature generator by combining the strength of a VAE and a GAN. In the current work, we use adversarial feature generation for memory replay in a continual learning framework. As demonstrated in~\cite{xian2018feature,xian2019f}, feature generation has achieved superior performance compared to image generation for zero-shot and few-shot learning.

% CCA Figure Find it in resources.tex
\ccafigure

\section{Forgetting in feature extractor and classifier}%\BR{I would simply call this section: Analyzing continual learning in classification networks}
\label{sec:analyzing}
In this section, we take a closer look at how forgetting occurs at different levels in a CNN.

\subsection{Class-incremental learning}

\minisection{Classification model and task.} We consider classification tasks learned from a dataset 
%$\mathcal{D}=\left\{\left(\mathbf{x}_i,\mathbf{y}_i\right)\right\}_{i=1}^{N}$
$\mathcal{D}=\left\{\left(\mathbf{x}_i,y_i\right)\right\}_{i=1}^{N}$, where $\mathbf{x}_i\in \mathcal{X}$ is the $i$th image, $y_i\in \mathcal{C}$ is the corresponding label (from a vocabulary of $K$ classes) %(as a one-hot representation with $K$ classes) 
and $N$ is the size of the dataset. The classifier network  has the form
% $\mathbf{\tilde y}=M\left(\mathbf{x}\right)=\mathcal{A}\left(H_{\phi}\left(F_{\theta}\left(\mathbf{x}\right)\right)\right)$
$\mathbf{\tilde y}=M\left(\mathbf{x};\theta,V\right)=H\left(F\left(\mathbf{x};\theta\right);V\right)$
, where we explicitly distinguish between \textit{feature extractor}  $F\left(\mathbf{x};\theta\right)$, parametrized by $\theta$, and \textit{classifier} $H\left(\mathbf{u}; V \right)=\mathcal{A}\left(V\mathbf{u}\right)$, where $V$ is a matrix projecting the output of the feature extractor $\mathbf{u}$ to the class scores (in the following we omit parameters $\theta$ and $V$), and $\mathcal{A}$ is the softmax function
% an activation function, typically a softmax
that normalizes the scores to class probabilities.
% , where each dimension estimates the class posterior $y_c=p\left(c\vert \mathbf{x}\right)$.
%$y}=\mathcal{A}_t\left(H_{\phi_t}\left(F_{\theta_t}\left(\mathbf{x}\right)\right)\right)$, where $F_{\phi_t}$ is the feature extractor, $H_{\theta_t}$
During training we minimize the cross-entropy loss between true labels and predictions $\mathcal{L}_{\textnormal{CE}}\left(\mathcal{D}\right)=-\Sigma_{i=1}^{N}\mathbf{y}_i \cdot \log \mathbf{\tilde y}_i$, where $\mathbf{y}_i$ is the one-hot representation of class label $y_i \in \mathcal{C}$.
%\left(\mathbf{x}_i,l_i\right)

\minisection{Continual learning.} 
% As in many previous works~[CITES], 
We consider the continual learning setting where $T$ classification tasks are learned independently and in sequence from the corresponding datasets $\mathcal{D}_1,\ldots,\mathcal{D}_t,\ldots,\mathcal{D}_T$. The resulting model $M_t$ after learning task $t$ has feature extractor $F_t$ and classifier $H_t$
% (and activation function $\mathcal{A}_t$, in case it is different)
. We assume that the classes in each task are disjoint, i.e. $\mathcal{C}_t \bigcap \mathcal{C}_{t'}=\emptyset$ for all $t'\neq t$. Ideally, after learning task $t$, the model can perform inference on all tasks $t'\leq t$ (i.e. it remembers current and previous tasks). We consider class-incremental learning in this work, where task-ID is unknown and  it requires predictions over all the classes learned so far. %, i.e. $\mathcal{C}'_t = \bigcup_{j=1}^t \mathcal{C}_j$. 

\subsection{Forgetting analysis of various methods}

\minisection{Fine-tuning.} In Figure~\ref{fig:feature_correlation_analysis} (far left) we illustrate the effect of continual learning (via simply fine-tuning the network on new tasks) on features extracted at different layers of the network. Forgetting is measured using Canonical Correlation Analysis (CCA) similarity\footnote{\label{foot:cca}CCA similarity computes the similarity between distributed representations even when they are not aligned. This is important, since learning new tasks may change how different patterns are distributed in the representation. We use SVCCA~\cite{raghu2017svcca} which first removes noise using singular value decomposition (SVD).} between the features extracted for task $t'\leq t$ by model $M_t$ and the optimal model $M_{t'}$ (i.e. trained at time $t'$ with $\mathcal{D}_{t'}$). Earlier features remain fairly correlated, while the correlation decreases progressively with increasing layer depth. This suggests that forgetting in higher-level features is more pronounced, since they become progressively more task-specific, while lower features are more generic.

\minisection{Learning without forgetting.}
A popular method to prevent forgetting is \textit{Learning without Forgetting} (LwF)~\cite{li2018learning}, which keeps a copy of the model $M_{t-1}$ before learning the new task and distills its predicted probabilities into the new model $M_{t}$ (which may otherwise suffer interference from the current task $t$). In particular, LwF uses a modified cross-entropy loss over each head of previous tasks given by
% \MM{Do we need to write the formula? If we lack space...}
$
% \begin{eqnarray}
    \mathcal{L}_{\textnormal{LwF}}\left(\mathcal{X}_t \right )=-\mathbb{E}_{\mathbf{x}\sim \mathcal{X}_t}\Sigma_{j=1}^{t-1}\tilde{\mathbf{y}}^{t-1,j} \cdot \log \tilde{\mathbf{y}}^{t,j}
    % \label{eq:lwf}
% \end{eqnarray}
$.

%are further renormalized from the output probabilities $\mathbf{y}$ as $\tilde{y}^{(c)}={\left(y^{(c)} \right )^{1/\mathcal{T}}}/{\Sigma^K_{k=1}\left(y^{(k)} \right )^{1/\mathcal{T}}}$ (with temperature $\mathcal{T}=2$). 

Note that the probabilities $\mathbf{\tilde y}^{t-1,j}$ and $\mathbf{\tilde y}^{t,j}$
% in equation~(\ref{eq:lwf}) 
are always estimated with current input samples $\mathbf{x}\in\mathcal{X}_t$, since data from previous tasks is not available. Since tasks are different, there is a distribution shift in the visual domain (i.e. $\mathbf{\tilde y}^{t-1,j}$ if extracted from $\mathbf{x}\in \mathcal{X}_{t-1}$ instead of $\mathbf{x}\in \mathcal{X}_t$), which can reduce the effectiveness of distillation when the domain shift from $\mathcal{X}_{t-1}$ to $\mathcal{X}_t$ is large. Figure~\ref{fig:feature_correlation_analysis} shows how LwF helps to increase the CCA similarity for previous tasks at the classifier, effectively alleviating forgetting and maintaining higher accuracy for previous tasks than fine tuning. However, the correlation at middle and lower-level layers in the feature extractor remains similar or lower to the case of fine tuning. This may be caused by the fact that the distillation constraint on the probabilities is too loose to enforce correlation in intermediate features.

\frameworkfigure % FIGURE FRAMEWORK

\minisection{Generative image replay.}
The lack of training images for previous tasks in continual learning has been addressed with a generator of images from previous tasks and using them during the training of current and future tasks~\cite{nguyen2017variational,ostapenko2019learning,shin2017continual,wu2018memory}. We consider conditional GAN with Projection Discriminator ~\cite{miyato2018cgans}, which can control the class of generated images. At time $t$, the image generator samples images $\hat{\mathbf{x}}=G_{t-1}\left(c,\mathbf{z}\right)$ where $c$ is the desired class and $\mathbf{z}$ is a random latent vector sampled from a simple distribution (typically a normalized Gaussian). These generated images are combined with current data in an augmented dataset $\mathcal{D}'_t=\{(\hat{\mathbf{x}}_i,{y}_i)\}_{i=1}^{N_R}\cup \mathcal{D}_t$, where $\hat{\mathbf{x}_i}=G_{t-1}({y}_i,\mathbf{z}_i)$ and $N_R$ is the number of replayed images for previous tasks (typically distributed uniformly across tasks and classes).

Generative image replay, while appealing, has numerous limitations in practice. First, real images are high dimensional representations and the image distribution of a particular task lies in a narrow yet very complex manifold. This complexity requires deep generators with many parameters and are computationally expensive, difficult to train, and often highly dependent on initialization~\cite{lucic2017gans}. Training these models requires large amounts of images, which is rarely the case in continual learning. Even with enough training images, the quality of the generated images is often unsatisfactory as training data for the classifier, since they may not capture relevant discriminative features. Figure~\ref{fig:feature_correlation_analysis} shows the CCA similarity for class-conditional GAN. It shows a similar pattern to LwF and fine tuning with the similarity decreasing especially in intermediate layers. 

\section{Feature distillation and generative feature replay}
\label{sec:distillation}
%Motivated by the previous observations, we propose generative \emph{feature} replay as an alternative to image replay. 
In the previous analysis of forgetting in neural networks, we saw that generative image  replay yields unsatisfactory results when applied to datasets that are difficult to generate (like CIFAR-100). We also observed that feature distillation prevents forgetting in the feature extractor. Therefore, to obtain the advantage of replay methods, which do not have the imbalance problem arising from multiple classification heads, we propose \emph{feature} replay as an alternative to image replay. We combine feature distillation and feature replay in a hybrid model that is effective and efficient. (see Figure~\ref{fig:motivation} right). 

Specifically, we use distillation at the output of the feature extractor in order to prevent forgetting in the feature extractor, and use feature replay of the same features to prevent forgetting in the classifier and to circumvent the classifier imbalance problem. Note that feature distillation has also been used in other applications~\cite{michieli2019incremental,tung2019similarity,yu2019learning}.

Our framework consists of three modules: feature extractor, classifier, and feature generator. To prevent forgetting we also keep a copy of the feature extractor, classifier and feature generator from the previous set of tasks. Figure~\ref{fig:framework} illustrates continual learning in our framework. The classifier $H_t$ and feature extractor $F_t$ for task $t$ are implicitly initialized with $H_{t-1}$ and $F_{t-1}$ (which we duplicate and freeze) and trained using feature replay and feature distillation. When the feature extractor and classifier are trained, we then freeze them and then train the feature generator $G_t$. A detailed algorithm is given in Algorithm~\ref{table:algorithm}.

\tablealgorithm  % Algorithm 

\subsection{Feature generator}
To prevent forgetting in the classifier we train a feature generator $G_t$ to model the conditional distribution of features $p_\mathbf{u}\left(\mathbf{u}\vert c\right)$ as 
$\hat{\mathbf{u}}=G_t\left(c, \mathbf{z}\right)$, and sample from it when learning future tasks. We consider two variants: \textit{Gaussian class prototypes}, \textit{conditional GAN with replay alignment}.
% and \textit{conditional GAN with classifier projection}.

\minisection{Gaussian class prototypes.}  We represent each class $c$ of a task $t$ as a simple Gaussian distribution $G_t(c,\mathbf{z})=\mathcal{N}(\mathbf{u} ; \mathbf{\mu}_t^{(c)},\mathbf{\Sigma}_t^{(c)})$, where $\mathcal{N}(\cdot; \cdot, \cdot)$ is a Gaussian distribution whose parameters are estimated using $\left\{\mathbf{u}_i=F_t\left(\mathbf{x}_i\right), \forall\left(\mathbf{x}_i,y_i\right)\in \mathcal{D}_t, y_i=c\right\}$. This variant has the advantage of compactness and efficient sampling.

\minisection{Conditional GAN with replay alignment.}
To generate more complex distributions and share parameters across classes and tasks, we propose to generate the feature extractor distribution with GANs. We use the Wasserstein GAN and adapt it to feature generation and continual learning using the following losses (between learning tasks $t$ and $t+1$):
\begin{eqnarray}
    \mathcal{L}_{D_t}^{\textnormal{WGAN}}\left(\mathcal{X}_t\right) &=& 
     +\mathbb{E}_{\mathbf{z}\sim p_{z},c \in C_t }\left [ D_t\left(c,G_t\left (c, \mathbf{z} \right )\right) \right ] \\ \nonumber
     & & -\mathbb{E}_{\mathbf{u}\sim \mathcal{D}_t}\left [ D_t\left(c,F_t\left(\mathbf{x}\right)\right) \right ] \\
    \mathcal{L}_{G_t}^{\textnormal{WGAN}}\left(\mathcal{X}_t\right) &=& -\mathbb{E}_{\mathbf{z}\sim p_{z},c\in C_t }\left[ D_t\left(c,G_t\left (c, \mathbf{z} \right )\right) \right].
\end{eqnarray}
%\begin{equation}
%         \mathcal{L}_{G_t}^{\textnormal{WGAN}}\left(\mathcal{X}_t\right) = -\mathbb{E}_{\mathbf{z}\sim %p_{z},c\in C_t }\left[ D_t\left(c,G_t\left (c, \mathbf{z} \right )\right) \right]
%\end{equation}
A replay alignment loss $\mathcal{L}_{G_t}^{\textnormal{RA}}$ is also added:
\begin{equation}
     \mathcal{L}_{G_t}^{\textnormal{RA}} = \Sigma_{j=1}^{t-1}\Sigma_{c\in C_j  }\mathbb{E}_{\mathbf{z}\sim p_\mathbf{z}}\left[\left\lVert G_t\left(c,\mathbf{z}\right) - G_{t-1}\left(c,\mathbf{z}\right)\right\rVert_2^2\right].
\end{equation}
which can be seen as a type of distillation~\cite{wu2018memory}. This replay alignment loss encourages the current generator $G_t$ to replay exactly the same features as $G_{t-1}$ when conditioned on a given previous class $c$ and a given latent vector $\mathbf{z}$. We use a discriminator $D_t$ during the adversarial training, which alternates updates of $D_t$ and $G_t$ (i.e. $\min_{D_t} \mathcal{L}_{D_t}^{\textnormal{WGAN}}\left(\mathcal{X}_t\right)$ and $\min_{G_t} \mathcal{L}_{G_t}^{\textnormal{WGAN}}\left(\mathcal{X}_t\right)+\mathcal{L}_{G_t}^{\textnormal{RA}}$, respectively).

\subsection{Feature extractor with feature distillation}
We prevent forgetting in $F_t$ by distilling the features extracted by $F_{t-1}$ via the following L2 loss:
\begin{eqnarray}
    \mathcal{L}_{F_t}^{\textnormal{FD}}\left(\mathcal{X}_t\right) = \mathbb{E}_{\mathbf{x}\sim \mathcal{X}_t}\left[\left\lVert F_t\left(\mathbf{x}\right) - F_{t-1}\left(\mathbf{x}\right)\right\rVert_2\right].
\end{eqnarray}
Note that there are no separate losses for each head (like in~\cite{li2018learning}) because the feature 
$\mathbf{u}=F\left(\mathbf{x}\right)$ is shared among tasks. Also, the loss can be applied on any feature (e.g. tensors). 
 Note in Fig.~\ref{fig:feature_correlation_analysis} (center) how the CCA similarity of our approach compared to LwF increases, which indicates that there is less forgetting. 

\subsection{Algorithm of class-incremental learning}

We are interested in a single head architecture that provides well-calibrated, task-agnostic predictions, which naturally arises if all tasks are learned jointly when all data is available. In our case we extend the last linear layer $V_{t-1}$ to $V_t$ by increasing its size to accommodate the new classes $C_t$. The softmax is also extended to this new size. During training we combine the available real data for the current task (fed to $F_t$) with generated features for previous tasks $\{(\hat{\mathbf{u}}_i,{y}_i)\}_{i=1}^{N_R}$. Since we only train a linear layer with features, this process is efficient.

Figure~\ref{fig:feature_correlation_analysis} (far right) shows that our method preserves similar representations for previous tasks at all layers, including the classifier. Our combination of distillation and replay maintains higher accuracy across all tasks, effectively addressing the problems of forgetting and task aggregation.

%%%%%%%%%%%%%%%%%%%%%%%%%%%%%%%%%%%%%%%%%%%%%%%%%%%%%%%%%%%%%%%%%%
%%%%%%%%%%%%%%%%%%%%%%%%%%% Experimental Results%%%%%%%%%%%%%%%%%%
%%%%%%%%%%%%%%%%%%%%%%%%%%%%%%%%%%%%%%%%%%%%%%%%%%%%%%%%%%%%%%%%%%

% ImageNet Figure
\imagenetsubfigure

\section{Experimental results}
\label{sec:experiments}
We report experiments evaluating the performance of our approach compared to baselines and the state-of-the-art. 

\minisection{Datasets.}
We evaluate performance on ImageNet~\cite{imagenet_cvpr09} and  CIFAR-100~\cite{krizhevsky2009learning}. ImageNet-Subset contains the first 100 classes in ImageNet in a fixed, random order. We resize ImageNet images to 256$\times$256, randomly sample 224$\times$224 crops during training, and use the center crop during testing. CIFAR-100 images are padded with 4 pixels, from which 32$\times$32 crops are randomly sampled. The original center crop is used for testing. Random horizontal flipping is used as data augmentation for both datasets.

%CUB-200-2011~\cite{WelinderEtal2010}

\minisection{Training.}
We use Pytorch as our framework ~\cite{paszke2017automatic}. For CIFAR-100, we modify the ResNet-18 network to use 3$\times$3 kernels for the first convolutional layer and train the model from scratch\footnote{This network setting was also used for the computation of Figure~\ref{fig:feature_correlation_analysis}.}. We train each classification task for 201 epochs and GANs for 501 epochs. For ImageNet, we use ResNet-18 and also train the model from scratch. We train  each classification task for 101 epochs and GANs for 201 epochs. The Adam optimizer is used in all experiments, and the learning rate for classification and GANs are 1e-3 and 1e-4, respectively. The classes for both datasets are arranged in a fixed random order as in~\cite{hou2019learning, rebuffi2017icarl}.  The coefficient of distillation loss is set to 1.

\minisection{Evaluation.}
The first evaluation metric is the average overall accuracy as in~\cite{hou2019learning, rebuffi2017icarl}. It is computed as the average accuracy of all tasks up to the current task. The second evaluation metric is the average forgetting measure as in~\cite{chaudhry2018riemannian}. It defines forgetting for a specific task as the difference between the maximum accuracy achieved on that task throughout the learning process and the accuracy the model currently achieves for it. The average forgetting is computed by averaging the forgetting for all tasks up to the current one. More evaluation metrics can be found in ~\cite{diaz2018don, lesort2020continual}

\subsection{Class-incremental learning experiments}

We first compare our approach with other methods on ImageNet-Subset and CIFAR-100.  We use half of the classes from each dataset as the first task and split the remaining classes in 5, 10 and 25 tasks with equally distributed classes (as also done in~\cite{hou2019learning}). In figures and tables ``Ours Gaussian'' indicates our method with Gaussian replay and ``Ours'' indicates our method with generative feature replay. We compare our approach with several methods: LwF~\cite{li2018learning}, EWC~\cite{kirkpatrick2017overcoming}, MAS~\cite{aljundi2018memory}, iCaRL~\cite{rebuffi2017icarl} and Rebalance~\cite{hou2019learning}. iCaRL-CNN uses a softmax classifier while iCaRL-NME uses the nearest mean classifier. The first three methods are trained without exemplars and iCaRL and Rebalance store 20 samples per class. For the first three methods, we train a multi-head network, where each task has a separate head since they will not work with single-head when there are no exemplars. We simply pick the maximum probability across all heads as the chosen output.

\cifarfigure
% Memory Usage table
\memorytable

\minisection{Comparative analysis on ImageNet-Subset.} We report the average accuracy and the average forgetting on ImageNet-Subset in Figure~\ref{fig:imagenet100_50}. It is clear that using exemplars for iCaRL and Rebalance is superior to most methods without exemplars, such as LwF, MAS and EWC. Our method with Gaussian replay performs similarly to iCaRL-NME and much better than iCaRL-NME in the 5 and 10 task setting. Surprisingly, it outperforms both iCaRL-CNN and iCaRL-NME by a large margin in the 25-task setting. By using GANs for replay, our method shows significant improvement compared to Gaussian replay and outperforms the state-of-the-art method Rebalance by a large margin. The gain increases with increasing number of tasks. It achieves the best results in all settings in terms of both average accuracy and forgetting. It is important to note that for our methods we do not need to store any exemplars from previous tasks and generated features are dynamically combined with current data. A comparison with other methods on ImageNet-1000 is in Appendix~\ref{image_full}. %\alert{Table~\ref{tabimagenet}}

\minisection{Comparative analysis on CIFAR-100.}
Results for CIFAR-100 are shown in Figure~\ref{fig:cifar100_50}. Our method with generative feature generation outperforms iCaRL, LwF, MAS and EWC by a large margin and achieves comparable results as Rebalance in the case of 5 and 10 tasks. We achieve slightly worse results in the 25-task setting compared to Rebalance, which might be because features from low resolution images are not as good as those learned from ImageNet. In contrast, for both iCaRL and Rebalance, 2000 exemplars in total must be stored. It is interesting that our method with Gaussian replay performs quite well compared to iCaRL, but slightly worse than Rebalance. 

\subsection{Comparison of storage requirements}
In Table~\ref{tab:memory} we compare the memory usage of exemplar-based methods iCaRL~\cite{rebuffi2017icarl} and Rebalance~\cite{hou2019learning}, the generative image replay method MeRGAN~\cite{wu2018memory}, and our generative feature replay. Exemplar methods normally store 20 images per class (from ImageNet or CIFAR-100), and the memory needed thus increases dramatically from 6.2MB to 375MB for 100 classes. Our approach, however, requires only a constant memory of 4.5MB for the generator and discriminator. For  256$\times$256$\times$3 images, our model is equivalent to only 24 total exemplars. Note that it is hard for exemplar-based methods to learn with only 24 exemplars. For larger numbers of classes such as full ImageNet-1000, it takes 3.8GB to store 20 samples per class. MeRGAN requires 8.5MB of memory, which is almost double the memory usage of ours. However, MeRGAN has difficulty generating realistic images for both CIFAR-100 and ImageNet and therefore obtains inferior results. %\XL{Training time ?}

\subsection{Generation of features at different levels} 
For our ablation study we use CIFAR-100 with 4 tasks with an equal number of classes. In Table~\ref{tab:cifar-feat} we look for the best depth of features to apply replay and distillation. We found that replaying at shallower depth results in dramatically lower performance. This is probably caused by: (1) the complexity of generating convolutional and lower-level features compared to the generation of linear high-level features from Block 4 (Ours); and (2) the difficulty of keeping the head parameters unbiased towards the last trained task when moving replay down in the network.

\ablationFeatureReplay

\section{Conclusions}
\label{sec:discussion}
We proposed a novel continual learning method that combines generative feature replay and feature distillation. We showed that it is computationally efficient and scalable to large datasets. Our analysis via CCA shows how catastrophic forgetting manifests at different layers. The strength of our approach relies on the fact that the distribution of high-level features is significantly simpler than the distribution at the pixel level and therefore can be effectively modeled with simpler generators and trained on limited samples. We perform experiments on the ImageNet and CIFAR-100 datasets. We outperform other methods without exemplars by a large margin. Notably, we also outperform storage-intensive methods based on exemplars in several settings, while the overhead of our feature generator is small compared to the storage requirements for exemplars. For future work, we are especially interested in extending the theory to feature replay for continual learning of embeddings~\cite{yu2020semantic}.

\minisection{Acknowledgement}
We acknowledge the support from Huawei Kirin Solution, the Industrial Doctorate Grant 2016 DI 039 of the Generalitat de Catalunya, the EU Project CybSpeed MSCA-RISE-2017-777720, EU's Horizon 2020 programme under the Marie Sklodowska-Curie grant agreement No.6655919 and the Spanish project RTI2018-102285-A-I00. 

{\small
\bibliographystyle{ieee_fullname}
\bibliography{egbib}
}

\clearpage
\begin{appendices}

\setcounter{figure}{5}
\begin{figure*}[tb]
	\centering
	\includegraphics[width=0.3\textwidth]{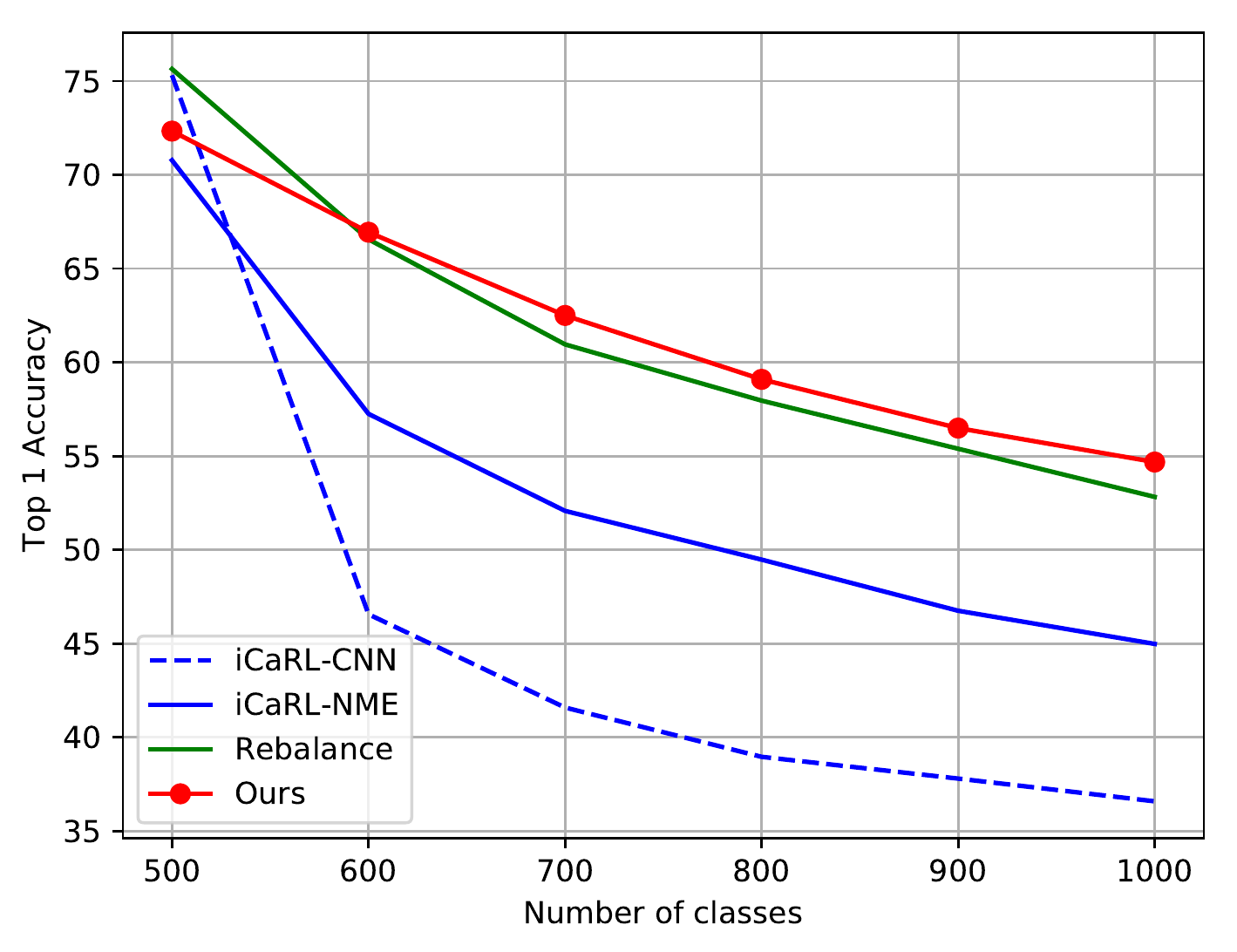}
	\includegraphics[width=0.3\textwidth]{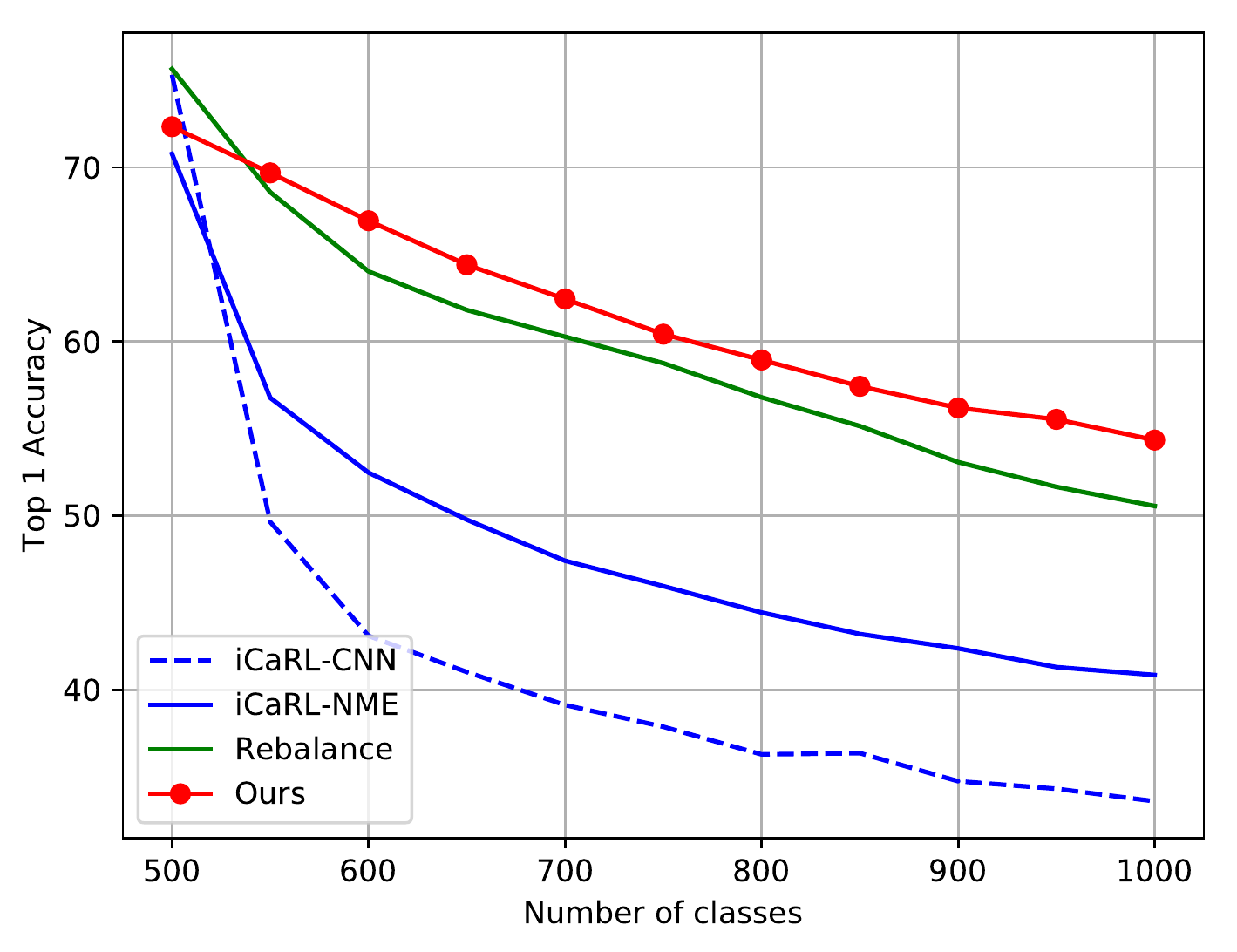}
	\includegraphics[width=0.3\textwidth]{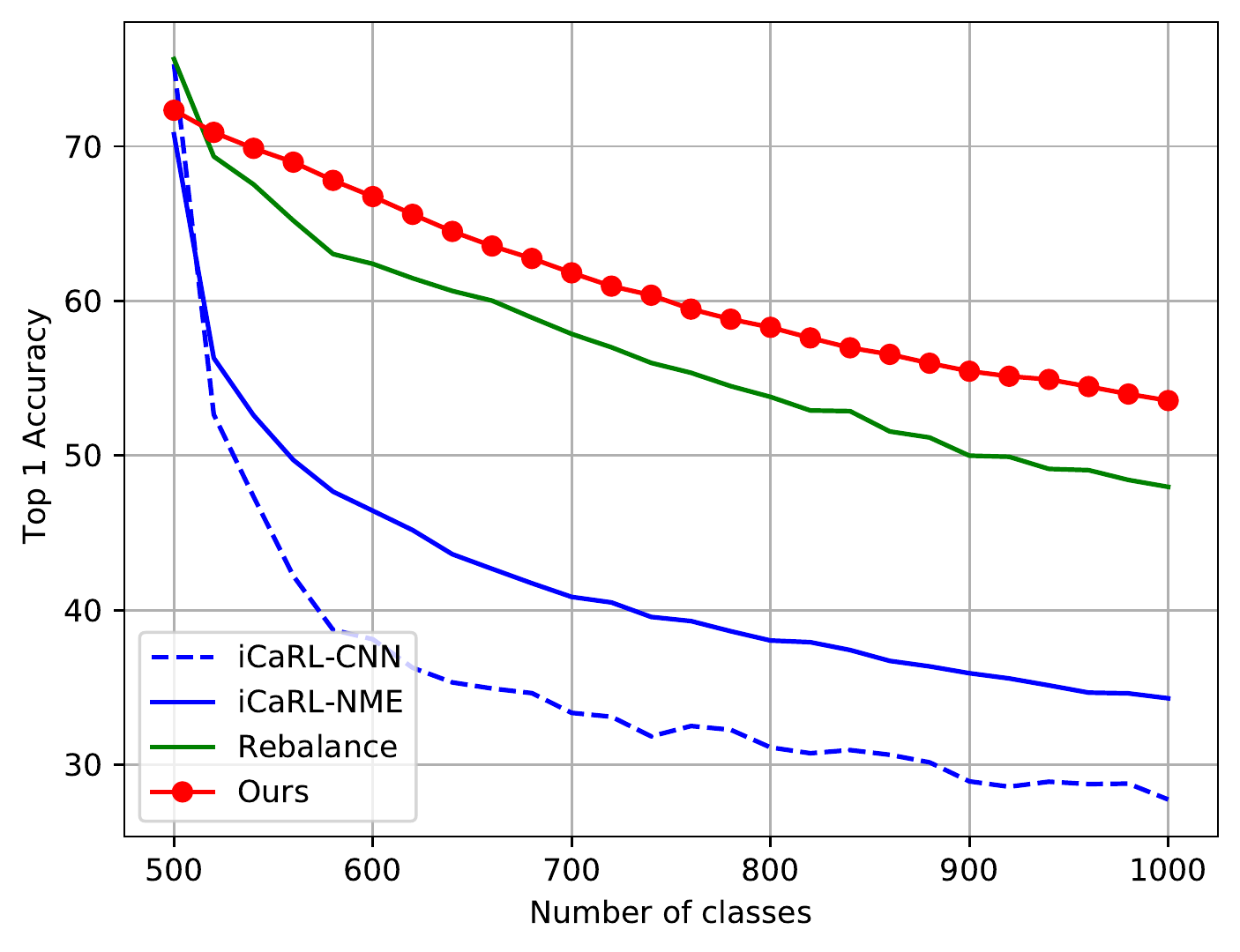}
	\includegraphics[width=0.3\textwidth]{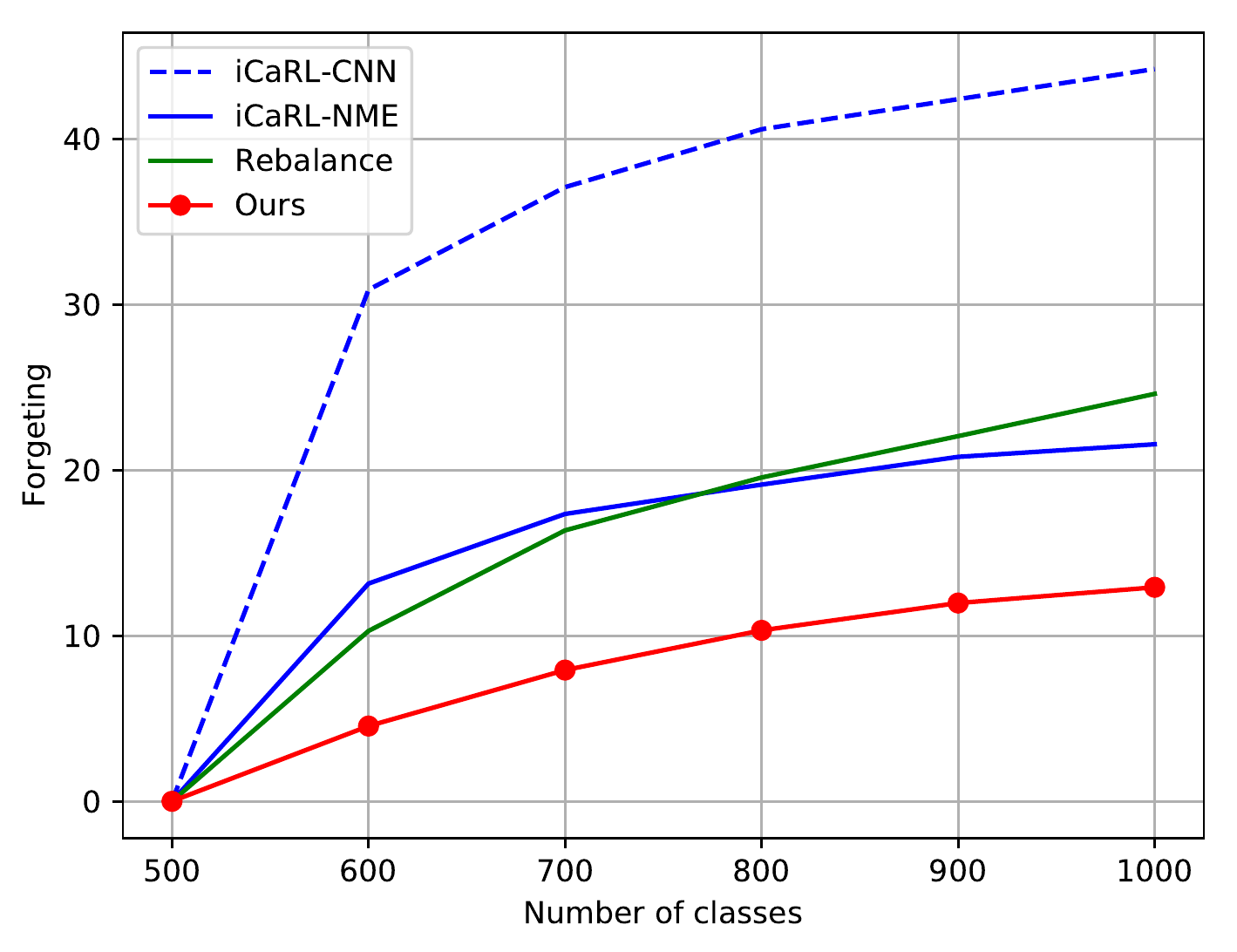}
	\includegraphics[width=0.3\textwidth]{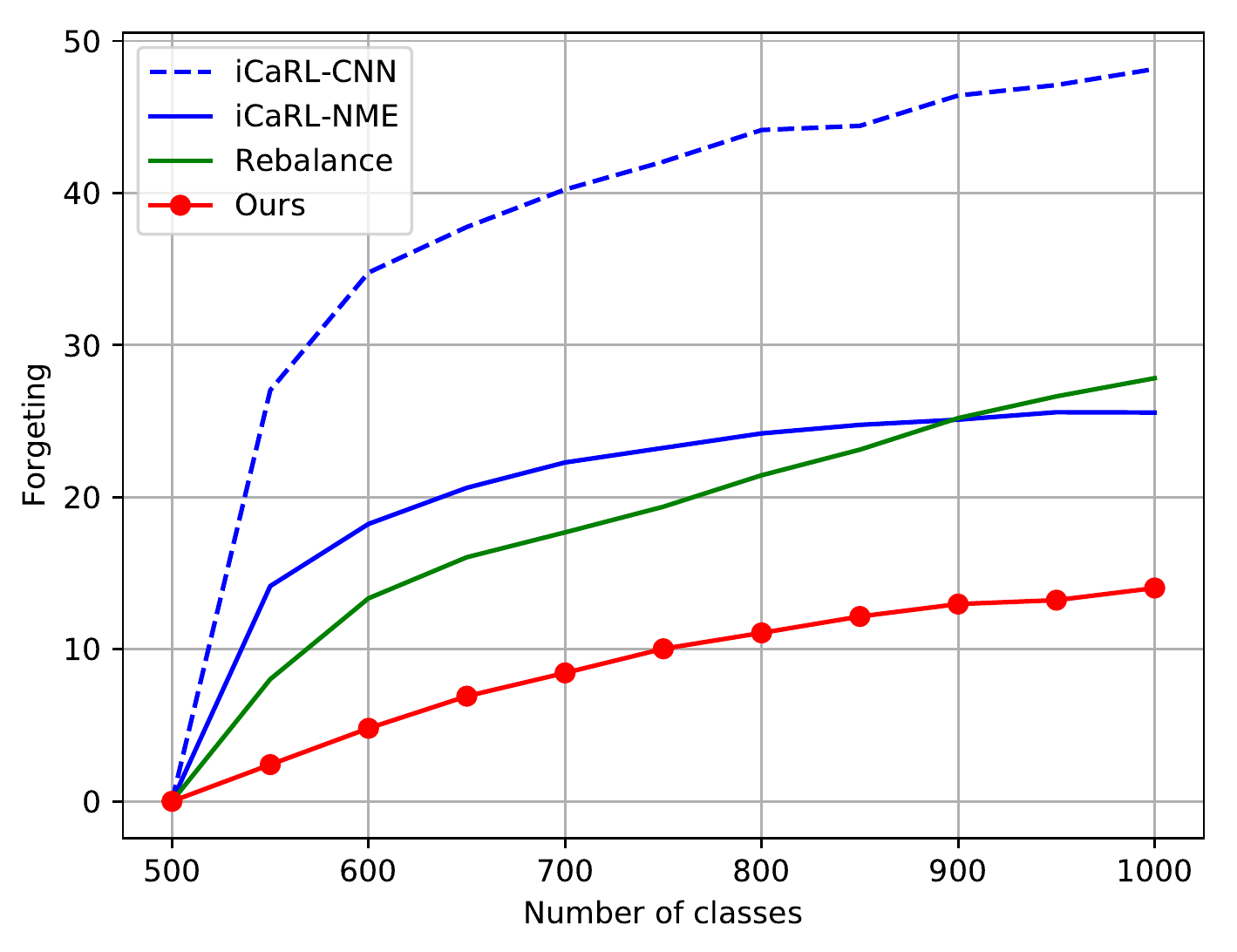}
	\includegraphics[width=0.3\textwidth]{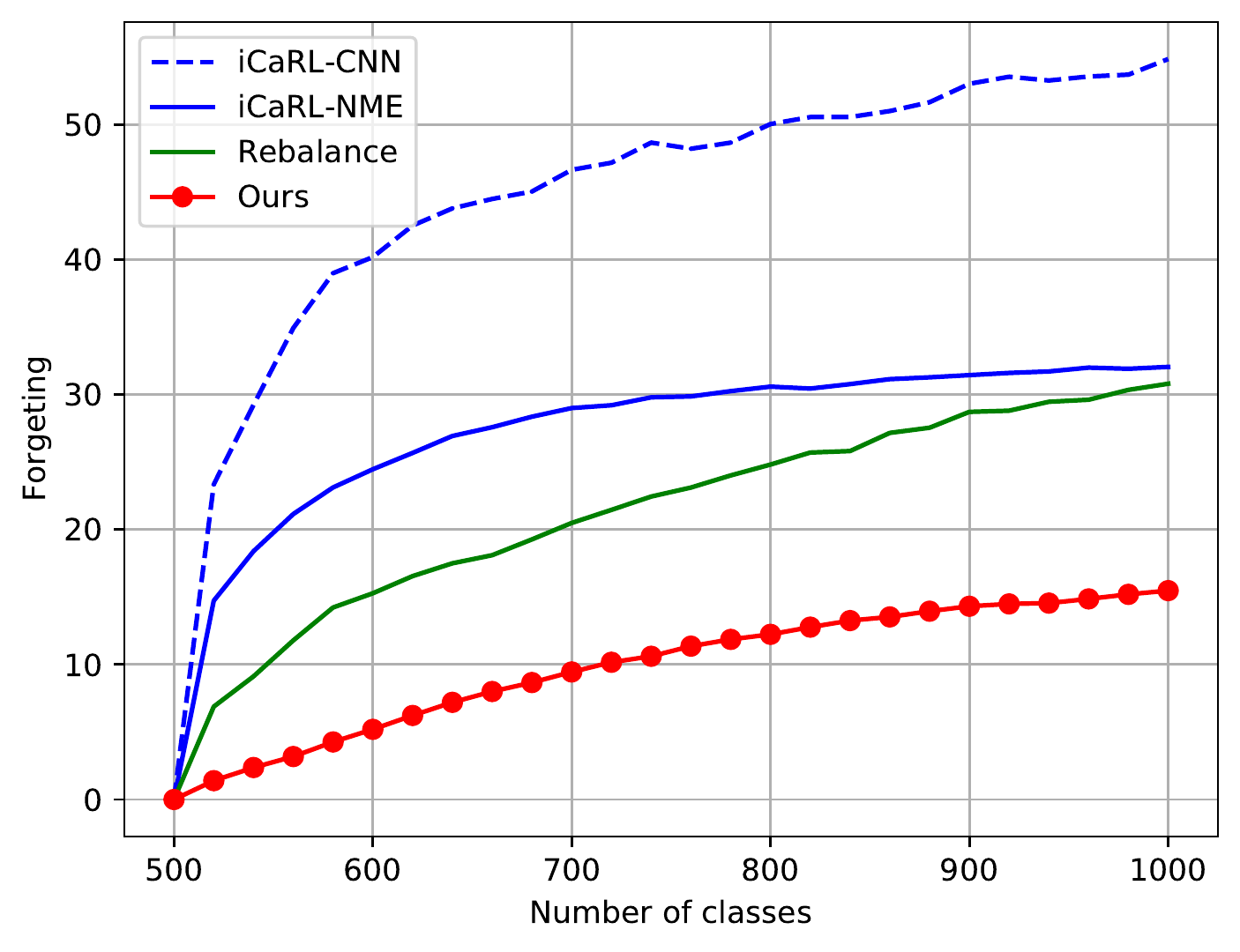}
	\caption{Comparison in the average accuracy (Top) and the average forgetting (Bottom) with various methods on ImageNet-1000. The first task has the half number of classes, and the remaining classes are divided into 5, 10 and 25 respectively. The lines with symbols are methods without using any exemplars, and without symbols are methods with 20000 exemplars.}
	\label{fig:imagenet_full}
\end{figure*}

\section{Comparative analysis on ImageNet-1000}
\label{image_full}
The average accuracy and forgetting on ImageNet-1000 are shown in Figure~\ref{fig:imagenet_full}. We can see that our proposed method outperforms iCaRL by a large margin in 5, 10 and 25 tasks. Compared to the state-of-the-art method Rebalance, we obtain slightly better accuracy in 5 tasks, and the gap is enlarged in both 10 and 25 tasks. In terms of the average forgetting, our method surpasses all methods by more than 10\%. It is important to note that for both iCaRL and Rebalance, they need to store 20000 exemplars in order to train in a continual setting. It takes about 3.8 Gb memory for these exemplar-based methods, while for our proposed method, we only need to store a generator and a discriminator with 4.5 Mb memory.

\section{Ablation study on different regularization}

For our ablation study we use CIFAR-100 with 4 tasks of equal number of classes. In Table~\ref{tab:cifar-reg} we compare different regularization methods in feature extractor, where feature distillation clearly outperforms MAS and EWC. This shows that adding constraints on features is superior to constraining in parameter space. This  guarantees that the generated features are closer to the real ones.

% Ablation study on regularization term
\setcounter{table}{2}
\ablationalignment

\section{T-SNE on generated features}

Here we show the T-SNE visualization of generated features using GANs and real features extracted from images (see Figure~\ref{fig:tsne}). We can see that the distributions of generated features and real features are very close, which allows our proposed method to train the classifier jointly with current data. There are clusters in the figures, which represents the distributions of different classes. 

\tsnefigure
\section{Architecture details}

Generator and Discriminator consist of two hidden layer of 512 neurons followed by LeakyReLU with parameter 0.2. We concatenate Gaussian noise $\mathbf{z}$ of 200 dimensions and one-hot vectors as input of Generator. More details can be seen in the available code.

\end{appendices}
\end{document}